\documentclass[letterpaper]{article} 

\usepackage{amsmath,amsfonts}

\def\eqref#1{equation~\ref{#1}}
\def\Eqref#1{Equation~\ref{#1}}

\def\1{\bm{1}}

\DeclareMathAlphabet{\mathsfit}{\encodingdefault}{\sfdefault}{m}{sl}
\SetMathAlphabet{\mathsfit}{bold}{\encodingdefault}{\sfdefault}{bx}{n}

\newcommand{\R}{\mathbb{R}}

\newcommand{\bfG}{{\bf G}}

\newcommand{\bfx}{{\bf x}}

\usepackage{aaai2026}  
\copyrighttext{Preprint — Not the camera-ready version.}

\makeatother

\usepackage{times}  
\usepackage{helvet}  
\usepackage{courier}  
\usepackage[hyphens]{url}  
\usepackage{graphicx} 
\usepackage{natbib}  
\usepackage{caption} 
\usepackage{algorithm}
\usepackage{algorithmic}

\makeatletter
\g@addto@macro{\UrlBreaks}{\do\_\do\/}
\makeatother
\usepackage{booktabs} 
\usepackage{multirow} 
\usepackage{amsmath}
\usepackage{amssymb}
\usepackage{amsthm}
\usepackage{cleveref}
\usepackage{enumitem}

\usepackage{newfloat}
\usepackage{listings}
\DeclareCaptionStyle{ruled}{labelfont=normalfont,labelsep=colon,strut=off} 
\floatstyle{ruled}
\newfloat{listing}{tb}{lst}{}
\floatname{listing}{Listing}
\title{Graph Flow Matching: Enhancing Image Generation with Neighbor-Aware Flow Fields}
\author {
    Md Shahriar Rahim Siddiqui\textsuperscript{\rm 1},
    Moshe Eliasof\textsuperscript{\rm 2},
    Eldad Haber\textsuperscript{\rm 1}
}
\affiliations {
    \textsuperscript{\rm 1}The University of British Columbia, Vancouver, Canada\\
    \textsuperscript{\rm 2}University of Cambridge, Cambridge, United Kingdom\\
    shahriar.siddiqui@ubc.ca, 
    me532@cam.ac.uk, ehaber@eoas.ubc.ca
}

\begin{document}

\maketitle

\begin{abstract}
Flow matching casts sample generation as learning a continuous-time velocity field that transports noise to data. Existing flow matching networks typically predict each point's velocity independently, considering only its location and time along its flow trajectory, and ignoring neighboring points. However, this pointwise approach may overlook correlations with points along the generation trajectory that could enhance velocity predictions, thereby improving downstream generation quality. To address this, we propose Graph Flow Matching (GFM), a lightweight enhancement that decomposes the learned velocity into a reaction term -- any standard flow matching network -- and a diffusion term that aggregates neighbor information via a graph neural module.  This reaction-diffusion formulation retains the scalability of deep flow models while enriching velocity predictions with local context, all at minimal additional computational cost. Operating in the latent space of a pretrained variational autoencoder, GFM consistently improves Fr\'echet Inception Distance (FID) and recall across five image generation benchmarks (LSUN Church, LSUN Bedroom, FFHQ, AFHQ-Cat, and CelebA-HQ at $256\times256$), demonstrating its effectiveness as a modular enhancement to existing flow matching architectures.
\end{abstract}


\section{Introduction}

Flow matching has recently gained traction as a promising generative modeling paradigm, framing sample synthesis as the integration of a learned (interpolated) continuous-time velocity field that deterministically or stochastically transforms noise into data \cite{lipman2023flowmatchinggenerativemodeling, chen2018neural}. Given two distributions $\pi_0$ and $\pi_1$, the task is to learn a velocity field $\mathbf{v}(\mathbf{x},t)$ that, when integrated, solves:
\begin{equation}
\label{eq:first_flow_equation}
\dot\bfx = \mathbf{v}(\mathbf{x}, t), \quad \mathbf{x}(0) \sim \pi_0, \quad \mathbf{x}(1) \sim \pi_1.
\end{equation}
We may view learning the field $\mathbf{v}(\mathbf{x},t)$ as an interpolation problem over the joint domain of $\mathbf{x}$ (namely $\mathcal{X} \subseteq \mathbb{R}^d$), and $t$ (namely $[0,1]$): $\mathcal X\times[0,1]$, where we supervise the model at discrete $(\mathbf{x},t)$ pairs sampled from a chosen transport plan, and require it to generalize to a velocity field everywhere in $\mathcal{X}$. Given $\mathbf{x}(0)$, which is typically a sample from a Gaussian distribution, the path in $\mathcal X\times[0,1]$ obtained by solving \Cref{eq:first_flow_equation} is what we will term a \emph{trajectory}. In the context of image generation, each $\mathbf{point}$ $\mathbf{x}(t)$ along such a trajectory is an image. 
While highly effective, current flow matching architectures predict each point's velocity \emph{pointwise}, conditioning only on its $(\mathbf{x}, t)$ coordinates without considering neighboring points in adjacent trajectories. This overlooks an important structural property shown in the literature \cite{gao2024gaussian, gao2024convergence, fukumizu2024flow} but not explicitly leveraged: the \emph{smoothness} of \(\mathbf{v}(\mathbf{x}, t)\) over \(\mathcal{X} \times [0,1]\). We observe that this regularity implies that neighboring points along nearby trajectories typically exhibit correlated behavior, which could be harnessed to improve flow estimation. 

\begin{figure*}[!t]
  \centering
  \includegraphics[scale=0.90]{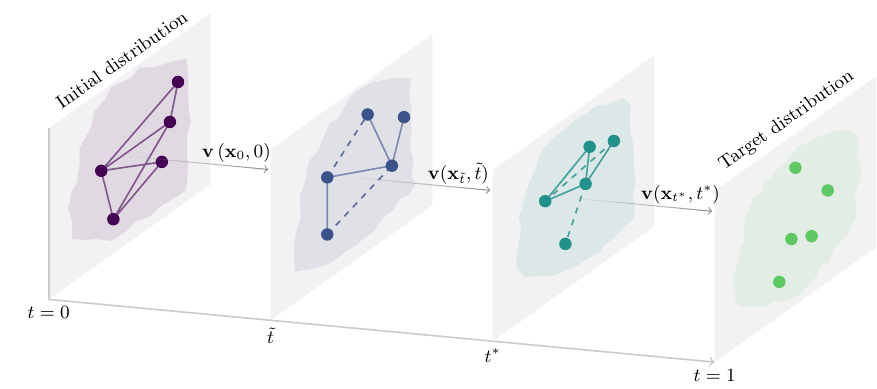}
  \caption{Graph flow matching enriches the flow trajectory from the initial distribution ($t=0$) to the target distribution ($t=1$) by connecting, at each intermediate time \(t\) (shown as slices) nodes $\mathbf{x}$ (shown as dots) that are \emph{latent vectors} (VAE codes) of distinct images using attention‑based similarity. The flow network output for each node $\mathbf{x}$ is its flow velocity $\mathbf{v}(\mathbf{x}, t)$. Each dot (node) at $t=0$ represents a Gaussian noise image, while each dot (node) at $t=1$ represents a generated image. Dashed edges indicate lower attention weights.}
  \label{fig:flow-matching}
\end{figure*}

In this paper, we leverage this observation, and propose \textbf{Graph Flow Matching (GFM)}, which enhances standard flow networks with neighbor awareness through a reaction-diffusion decomposition:
\begin{equation}
\mathbf{v}(\mathbf{x}, t) = \mathbf{v}_{\text{react}}(\mathbf{x}, t) + \mathbf{v}_{\text{diff}}(\mathbf{x}, t;\, \mathcal{N}(\mathbf{x}, t)),
\end{equation}
where $\mathbf{v}_{\text{react}}$ is any standard flow network, and $\mathbf{v}_{\text{diff}}$ is a lightweight graph-based correction term (a graph neural network) that aggregates information from the set of neighbors of $\mathbf{x}$ at time $t$, denoted by $\mathcal{N}(\mathbf{x},t)$. Figure~\ref{fig:flow-matching} illustrates this process, where the flow trajectory from $\pi_{0}$ to $\pi_{1}$ is augmented by incorporating neighborhood structure via graphs at intermediate time steps. Designing a flow matching pipeline requires two key decisions: \emph{network architecture} and \emph{training strategy} (e.g., Rectified Flow \cite{RectifiedFlow_liu2022flow}, Consistency FM \cite{ConsistencyFM_yang2024consistencyfm}). We focus on the \emph{architecture} aspect, proposing an enhancement compatible with any training strategy and existing base flow network $\mathbf{v}_{\text{react}}$, pretrained or otherwise.

In this work, we adopt the latent-space generative modeling paradigm, popularized by work such as Stable Diffusion~\cite{StableDiffusion_rombach2022high}, and recently used for flow matching as in LFM~\cite{dao2023flow}, where image generation is performed in a perceptually compressed latent space produced by a pretrained variational autoencoder (VAE) \cite{VAE_kingma2022autoencodingvariationalbayes}. This choice reduces computational cost as the latent space defines meaningful distance metrics for constructing graphs, which are easy to compute. Our contributions therefore operate entirely within this latent framework. Therefore, in this work, each “point” $\mathbf{x}$ is the VAE latent encoding $\mathbf{x}\in\R^{4\times32\times32}$ of an image (in this paper, the latent encoding of a $256 \times 256$ RGB image); all graphs operate over these latent encodings (otherwise known as \emph{latent codes}). We shall denote these latent codes as $\mathbf{x}$ in this paper. 
Figure \ref{fig:ADM_FFHQ_comparisons_teaser} and Figure \ref{fig:lsun_bed_church_comparisons} (in the Appendix) qualitatively demonstrate our method's effectiveness across three diverse datasets. Quantitatively, we validate GFM on five unconditional image generation benchmarks—LSUN Church \cite{LSUN_dataset_yu15lsun}, LSUN Bedroom \cite{LSUN_dataset_yu15lsun}, FFHQ \cite{FFHQ_dataset_karras2019style}, AFHQ-Cat \cite{AFHQ_dataset_choi2020starganv2}, and CelebA-HQ \cite{CelebAHQ_dataset_karras2018progressive} at $256\times 256$ resolution, showing consistent improvements in Fr\'echet Inception Distance (FID) and recall while adding minimal computational overhead ($\lesssim 10\%$ additional parameters).

\paragraph{Our Contributions} (i) We introduce Graph Flow Matching (GFM), a modular reaction–diffusion framework that enhances flow networks with neighbor-aware velocity correction. GFM decomposes the velocity field into a standard pointwise term (using any existing flow network) and a graph-based diffusion term, integrating seamlessly with existing backbones and training strategies without requiring modifications to losses or solvers; (ii) We validate GFM in the latent space of a pretrained VAE, demonstrating consistent improvements in unconditional image generation across five standard benchmarks. We use two architectures for $\mathbf{v}_{\text{diff}}$, which we call MPNN and GPS, to empirically validate that the use of a graph module in the flow model this way is a promising enhancement to existing flow architectures.

\begin{figure*}[t]
  \centering
  \includegraphics[width=0.90\textwidth]{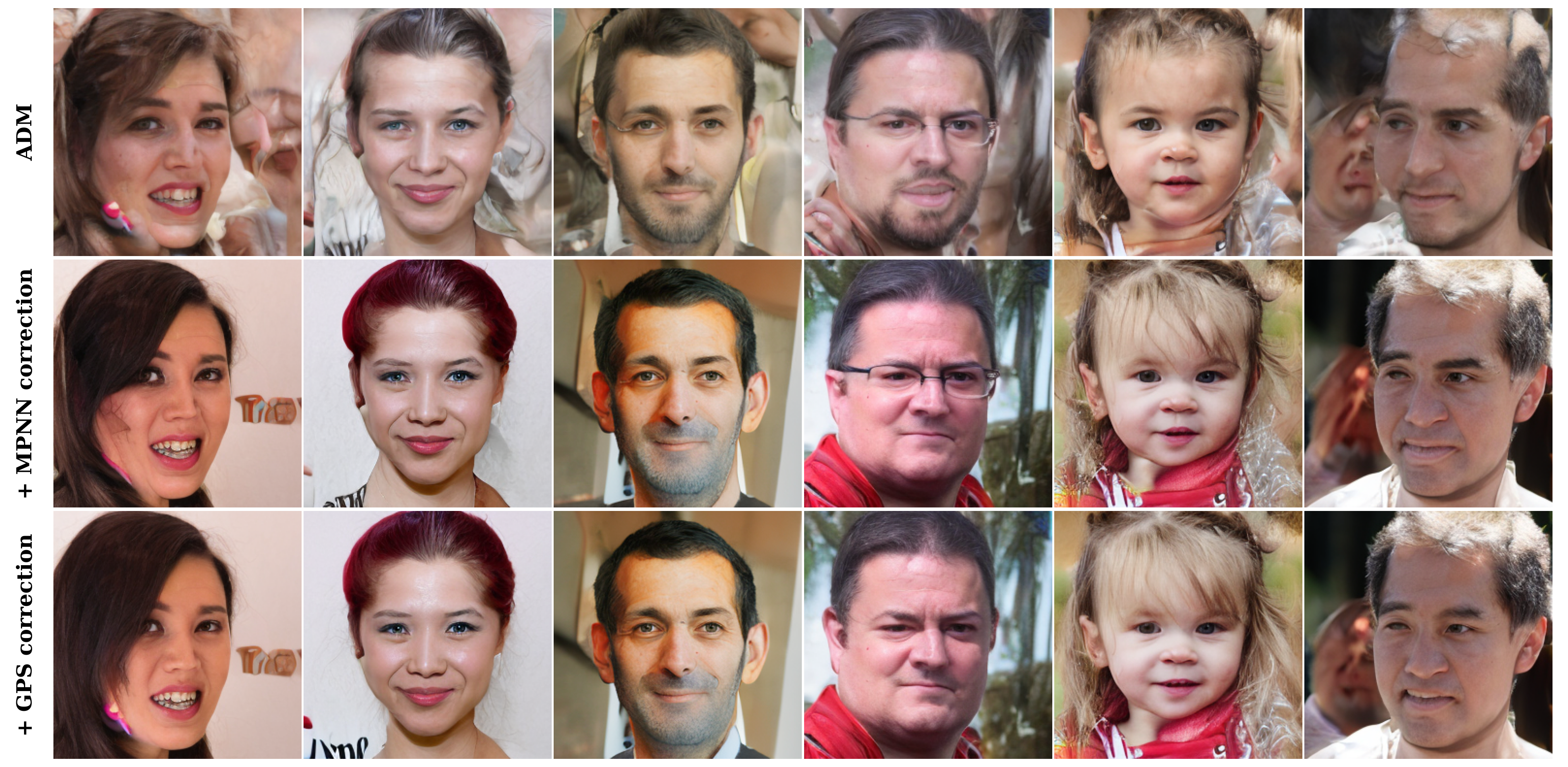}
  \caption{
  \textbf{Neighbor-aware flow matching enhances image generation.} FFHQ samples ($256\times256$) generated using the same random seed by: \textbf{(top)} baseline ADM U-Net \cite{dao2023flow}, \textbf{(middle)} ADM with MPNN-based correction, and \textbf{(bottom)} ADM with a GPS-based correction module \cite{GPS_rampavsek2022recipe}. GFM variants produce more coherent facial features and sharper details compared to the baseline model.
  }
  \label{fig:ADM_FFHQ_comparisons_teaser}
\end{figure*}

\section{Background and Related Work}
\label{sec:background}
A central challenge in generative modeling is transporting a simple base distribution (e.g., Gaussian noise) into a complex, high-dimensional target distribution. This underpins models such as generative adversarial networks (GANs)~\cite{GAN_goodfellow2014generative}, variational autoencoders (VAEs)~\cite{VAE_kingma2022autoencodingvariationalbayes}, normalizing flows~\cite{papamakarios_NormalizingFlows}, and diffusion models~\cite{song2019generative, song2021scorebasedgenerativemodelingstochastic, DDPM_ho2020denoising}.
Recent trends increasingly adopt continuous-time formulations. Continuous normalizing flows (CNFs)~\cite{chen2018neural} for example, generalize discrete-time flows by learning transformations via ODEs. Diffusion models, originally discrete-time~\cite{DDPM_ho2020denoising}, have evolved into continuous-time score-based models defined through stochastic differential equations (SDEs)~\cite{song2021scorebasedgenerativemodelingstochastic}.
Flow matching~\cite{lipman2023flowmatchinggenerativemodeling} offers a deterministic, continuous-time alternative that avoids score estimation by directly supervising a velocity field that transports noise samples (that is, samples from a Gaussian distribution) to data. It combines fast sampling and simplified training, making it an attractive framework for generative modeling. Our work builds upon this foundation by introducing a graph-based architectural enhancement that enriches velocity predictions with information from neighboring points along the flow trajectory. 
On a different note, while recent works use flow matching \emph{for} graph generation~\cite{VarFM_eijkelboom2024variational, song2023equivariant_3DMoleculeGeneration, Defog_qin2024defog, GraphCondFlowMatching_scassola2025graph, LatentSpaceGraphDiffusion_pombala2025exploring}, our GFM approach focuses on a different goal; it uses graph learning techniques \emph{for} enhancing flow matching techniques. 

\subsection{Flow Matching Preliminaries}

Flow matching formulates sample generation as solving an ordinary differential equation (ODE) that transports a source distribution $\pi_0$ to a target distribution $\pi_1$ over a finite time horizon $t \in [0,1]$. Given two random variables $X_0 \sim \pi_0$ and $X_1 \sim \pi_1$ with support in $\mathcal{X} \subseteq \mathbb{R}^d$, the goal is to learn a velocity field $\mathbf{v}_\theta : \mathbb{R}^d \times [0,1] \to \mathbb{R}^d$ such that the solution $\mathbf{x}(t)$ to the initial value problem
\begin{equation}
\label{eq:ode-flow}
\dot\bfx= \mathbf{v}_\theta(\mathbf{x}, t), \quad \mathbf{x}(0) \sim \pi_0,
\end{equation}
satisfies $\mathbf{x}(1) \sim \pi_1$. Learning the velocity field from data fundamentally constitutes an interpolation problem over the joint domain $\mathcal{X} \times [0,1]$. Given velocity supervision at a finite set of spatiotemporal points $(\mathbf{x}_{t_{i}}, t_i)$, the model must generalize to unseen points throughout this domain. Flow matching methods create synthetic supervision by sampling pairs $(\mathbf{x}_0, \mathbf{x}_1) \sim (\pi_0, \pi_1)$, selecting an interpolation time $t \in [0,1]$, constructing an interpolation point $\mathbf{x}_t$ between $\mathbf{x}_0$ and $\mathbf{x}_1$, and defining a target velocity $\mathbf{v}^*(\mathbf{x}_t, t)$ according to a transport plan. For example, under constant-velocity transport:
\begin{equation}
\label{eq:velocity-supervision}
\mathbf{v}^*(\mathbf{x}_t, t) = \mathbf{x}_1 - \mathbf{x}_0, \quad \mathbf{x}_t = (1-t)\mathbf{x}_0 + t \mathbf{x}_1.
\end{equation}

The training objective minimizes the squared error between predicted and target velocities:
\begin{equation}
\label{eq:flow-loss}
\mathcal{L}_{\text{FM}}(\theta) = \mathbb{E}_{t,\mathbf{x}_0,\mathbf{x}_1}\left[\left\|\mathbf{v}_\theta\left(\mathbf{x}_{t}, t\right) - \mathbf{v}^*(\mathbf{x}_{t},t)\right\|^2\right].
\end{equation}

This approach transforms sample generation into a supervised learning problem over velocity samples $(\mathbf{x}_t, t, \mathbf{v^*})$ drawn from the joint space $\mathcal{X} \times [0,1]$. At inference time, one generates new samples by numerically solving the learned ODE \Eqref{eq:ode-flow} starting from $X_0 \sim \pi_0$.

\subsection{Interpolation Perspective}
The flow velocity field \(\mathbf{v}(\mathbf{x}_t,t)\) is smooth, typically Lipschitz continuous in \(\mathbf{x}\) and \(t\) \cite{gao2024gaussian, gao2024convergence}. Moreover, the learned field satisfies the continuity equation, ensuring conservative mass transport \cite{albergo2022building, lipman2023flowmatchinggenerativemodeling}. Learning \(\mathbf{v}\) from a finite set of samples thus amounts to a scattered‐data interpolation problem: one observes velocities at sparse \((\mathbf{x}_t, t)\) locations and seeks to reconstruct a smooth vector field throughout the domain. Classical interpolants, such as radial basis functions, splines, and moving least squares, recover smooth functions by fitting to nearby samples with local support \cite{micchelli1984interpolation,de1978practicalSplines,levin1998_MLSapproximation}. 

By contrast, most flow networks predict each \(\mathbf{v}(\mathbf{x}_t,t)\) independently, ignoring correlations among neighboring points along adjacent flow trajectories. To explicitly instill this inductive bias into the architecture, we build an attention‐weighted graph over the batch at each timestep and use a lightweight GNN to allow each velocity prediction to draw on its neighbors, mimicking how interpolation stencils blend local information. We view the GNN term, \(\mathbf{v}_{\text{diff}}\), as a learnable analogue to classical interpolation kernels, generalizing fixed support functions to trainable networks. Further discussion is included in \Cref{app:FromInterpToNeighborFlowPred}.

\subsection{Graph-based Approaches in Deep Learning}

The limitations of pointwise processing have motivated significant research into graph-based architectures that explicitly model relationships between data points. Graph Neural Networks (GNNs) have emerged as a powerful framework for incorporating local structure, enabling each point to aggregate information from its neighbors through message passing mechanisms \cite{GCN_paper_kipf2017semi, GAT_paper_velickovic2018graph}.

In geometric deep learning, Graph Convolutional Networks (GCNs) \cite{GCN_paper_kipf2017semi}, Graph Attention Networks (GATs) \cite{GAT_paper_velickovic2018graph}, and Message Passing Neural Networks (MPNNs) \cite{MPNN_paper_gilmer2017neural} have demonstrated the value of neighborhood aggregation across diverse domains. These architectures naturally encode inductive biases about local structure, leading to improved generalization in tasks ranging from molecular property prediction to social network analysis.

Recent work has also explored the connection between GNNs and partial differential equations (PDEs). \cite{PDE_GCN_NEURIPS2021_1f9f9d8f} showed that the dynamics of PDEs can be encoded into GNN architectures, with the implicit PDE structure greatly influencing their suitability for specific tasks. This PDE-GNN connection provides a principled framework for designing graph architectures that capture desired physical dynamics, which is a key inspiration for the $\textbf{MPNN}$ graph correction module used in our experiments, as outlined in \Cref{sec:experiments}.

\subsection{Physical Inspiration: Reaction-Diffusion Systems}

Reaction-diffusion systems \citep{turing1990chemical}, which model how substances spread and interact in space, provide a natural framework for incorporating local interactions into flow matching. In these systems, the evolution of a quantity $u(\mathbf{x}, t)$ is governed by:
\begin{equation}
\label{eq:physical_reaction_diffusion_eq}
\frac{\partial u}{\partial t} = D \Delta u + R(u),
\end{equation}
where $D \Delta u$ represents diffusion and $R(u)$ represents reaction (pointwise dynamics).

This decomposition inspires our approach: by treating the standard flow matching velocity network as a reaction term that captures pointwise dynamics and adding a graph based diffusion term that allows neighbor to neighbor communication, we enable velocity predictions to adapt based on local flow patterns. 
In particular, the reaction-diffusion equation is known to generate complex patterns \cite{turing1990chemical} that can explain many natural processes.  Previous work has successfully utilized this equation in the context of GNNs \cite{haber2019imexnet,choi2023gread, eliasof2024graph} for solving graph related problems.

\subsection{Latent-Space Generative Modeling}
\label{sec:latent-space-modeling}

Generating high-resolution images directly in pixel space is computationally expensive and often burdens generative models with the need to reconstruct low-level details that are perceptually insignificant. Latent-space generative modeling mitigates this challenge by performing generation in a lower-dimensional, semantically meaningful space. This approach, popularized by models like Stable Diffusion~\citep{StableDiffusion_rombach2022high}, uses a two-stage process: a VAE first compresses images into latent representations, and a generative model then operates within this latent space to model data distributions.

Formally, given a pre-trained VAE with encoder $E: \mathcal{I} \rightarrow \mathcal{X}$ and decoder $D: \mathcal{X} \rightarrow \mathcal{I}$, latent-space generative modeling learns to transform a  prior distribution 
(e.g., Gaussian noise) 
into the distribution of encoded data points in $\mathcal{X}$. Here \(\displaystyle \mathcal{I} = [-1,1]^{H\times W\times C}\) denotes the pixel-space of \([-1,1]\)-normalized \(H\times W\times C\) images (as used by the Stable Diffusion VAE), and \(\mathcal{X}\subseteq \mathbb{R}^d\) its latent representation. The decoder $D$ then reconstructs image samples from this latent trajectory in $\mathcal{X}$ to $\mathcal{I}$. This formulation provides several critical advantages:
\begin{itemize}[leftmargin=3em]
    \item \textbf{Dimensionality Reduction.} Working in latent space reduces spatial resolution and channel complexity (e.g., from $256 \times 256 \times 3$ images to $32 \times 32 \times 4$ latent tensors), significantly lowering memory and compute requirements.
    \item \textbf{Semantic Compression.} The VAE, trained with perceptual and reconstruction losses, discards high-frequency noise and retains structurally and semantically relevant content. This reduces the burden on the generative model to learn low-level detail.
    \item \textbf{Decoupled Design.} Generation and reconstruction are modular: the VAE handles image fidelity, while the generative model focuses solely on learning distributional transformations in latent space.
\end{itemize}

Building on this design, the \emph{Latent Flow Matching (LFM)} framework~\citep{dao2023flow} extends flow matching to this perceptually aligned latent space. Rather than learning stochastic score-based dynamics as in diffusion models, LFM learns a deterministic velocity field that transports noise into encoded data samples by solving an ODE in latent space. This enables faster inference, fewer function evaluations, and simplified training compared to pixel-space diffusion.

For GFM to be effective, it must connect "neighboring" points to a graph. An embedded space is, therefore, natural to us as it defines a distance in latent space, which is easy to compute and is meaningful for complex data. Therefore, we build GFM  directly on this latent space setting. We adopt the same VAE architecture as LFM—namely, the Stable Diffusion VAE—which maps $256 \times 256$ RGB images into $32 \times 32 \times 4$ latent tensors. We retain all components of the LFM pipeline, including the constant-velocity transport plan, flow matching loss, and ODE solver. Our contribution lies exclusively in augmenting the velocity field $\mathbf{v}_\theta(\mathbf{x}, t)$ with a neighbor-aware graph based correction term. While there are many possible architectures for this part, here, we experimented with two architectures discussed below (\Cref{sec:method}).

\section{Graph Flow Matching}
\label{sec:method}
\vspace{2pt}
We propose \textit{Graph Flow Matching (GFM)}, a modular enhancement to flow matching networks that integrates local neighborhood structure through a lightweight graph module. GFM operates under a reaction–diffusion framework and is model-agnostic, where the velocity field is decomposed into a standard reaction term which could be any off-the-shelf pointwise acting flow network, and a graph-based message passing term (the so-called "diffusion" term that we propose). 

\subsection{Reaction–Diffusion Decomposition}
Given interpolation triplets $(\mathbf{x}, t, \mathbf{v})$ constructed via a transport plan during the training phase, GFM predicts the velocity at each point as:
\begin{equation}
\mathbf{v}_\theta(\mathbf{x}, t) = \mathbf{v}_{\text{react}}(\mathbf{x}, t) + \mathbf{v}_{\text{diff}}(\mathbf{x}, t;\, \mathcal{N}(\mathbf{x}, t)),
\label{eq:reaction-diffusion-decomp}
\end{equation}
where $\mathbf{v}_{\text{react}}$ is any off-the-shelf flow matching architecture, and $\mathbf{v}_{\text{diff}}$ is a graph-based correction term informed by a neighborhood $\mathcal{N}(\mathbf{x}, t)$ of surrounding samples at time $t$.

\textbf{Graph Generation.} 
At each intermediate timestep \(t\), we treat the VAE latent code \(\mathbf{x}_t \in \mathbb{R}^{4\times32\times32}\) of every sample in the minibatch as a graph node. We compute pairwise attention scores between these nodes. The attention weights form the adjacency matrix \(\mathbf{A}\), so that \(\mathcal{N}(\mathbf{x}_t,t)=\{\mathbf{x}_j\mid \mathbf{A}_{ij}>0\}\) is exactly the set of latent codes connected to $\mathbf{x}_{t}$, with node index \(i\) in the graph. In \Cref{sec:app_knn_ablation} we provide an ablation where the K-nearest-neighbors with learnable weights was used to construct the graph.

\textbf{Reaction Term: Standard Flow Networks.} 
The reaction term $v_{\text{react}}(\mathbf{x}, t)$ can be any off-the-shelf flow matching network that predicts velocity from spatial and temporal coordinates. It typically takes the form of a U-Net~\citep{UNet_ronneberger2015u, Dhariwal_ADM_NEURIPS2021_49ad23d1}, Vision Transformer~\citep{VisionTransformer_dosovitskiy2020image, DiT_peebles2023scalable}, or any pointwise neural interpolator. This component models pointwise transport based on global training dynamics.

In our experiments (Sec.~\ref{sec:experiments}), we instantiate $v_{\text{react}}$ with a UNet (ADM) \cite{Dhariwal_ADM_NEURIPS2021_49ad23d1} and a transformer model (DiT) \cite{DiT_peebles2023scalable}, following the settings in \citep{dao2023flow}. However, GFM is compatible with any flow matching backbone and training strategy.

\textbf{Diffusion Term: Neighbor-Aware Velocity Correction.}
\label{sec:DiffusionTermsUsed}
While the diffusion term $\mathbf{v}_{\text{diff}}(\mathbf{x}, t; \mathcal{N}(\mathbf{x}, t))$ can be implemented 
using any graph-based neural architecture that aggregates information from neighboring samples, here we focus on two such architectures: the Message Passing Neural Network (MPNN) \cite{MPNN_paper_gilmer2017neural} and a graph transformer architecture (GPS) \cite{GPS_rampavsek2022recipe}.  

\textbf{MPNN Architecture.}
The first diffusion term we used is a custom MPNN architecture where we use a graph gradient $\bfG$ (sometimes referred to as an incidence matrix) \cite{PDE_GCN_NEURIPS2021_1f9f9d8f} to compute the difference between node features. This yields an edge-based quantity, or edge features. We then apply a nonlinearity to the edge features and aggregate them back to the node using the transpose of the incidence matrix. We then apply a non-linear network to the result. The network (\Cref{eq:reaction-diffusion-decomp}) can thus be summarized by the following ODE
\begin{equation}
\label{eq:MPNN}
\begin{split}
  \mathbf{v}_\theta(\mathbf{x}_t,t) &= \frac{d \mathbf{x}_t}{dt} \\[-0.3ex]
  &= -\,N^{(2)}_{\!\theta}\bigl(G^\top[\sigma(G\,N^{(1)}_{\!\theta}(\mathbf{x}_t,t))],\,t\bigr) 
     + R(\mathbf{x}_t,t)\,.
\end{split}
\end{equation}
where \( R(\mathbf{x}_t, t) \) denotes the reaction term \( v_{\text{react}} \), \(\sigma \) denotes a nonlinearity and $N_{\theta}^{(1)}$ and $N_{\theta}^{(2)}$ are lightweight convolutional networks. This network generalizes the classical MPNN that uses Multi-Layer Perceptrons (MLPs) for $N_{\theta}^{(1)}$ and $N_{\theta}^{(2)}$. While the node data for classical graphs is unstructured, in our experiments, each node in the graphs represents an image. Hence, we utilize convolutional architectures (such as UNets) for $N_{\theta}^{(1)}$ and $N_{\theta}^{(2)}$ in this paper.

\textbf{GPS Architecture.}
Our second instantiation of the graph correction module employs the General, Powerful, Scalable (GPS) graph transformer architecture~\cite{GPS_rampavsek2022recipe}.
We adapt the GPS framework for flow matching as follows: (i) recasting each latent tensor in the batch as a node in a fully connected graph; (ii) integrating temporal information via learned time embeddings that are projected and concatenated with node features; (iii) incorporating random walk positional encodings (RWPE) generated at runtime using Pytorch Geometrics's implementation; and (iv) replacing the standard GINEConv blocks with recurrent GatedGraphConv units. The GPS architecture alternates local message passing and global multi-head attention, allowing nodes to aggregate information from the entire batch. Finally, the network projects the enriched node representations back to the original latent tensor dimensions, producing the diffusion velocity correction term. 
The adjacency matrix of this setup is thus attention-based.

\textbf{Complexity.} 
Graph Flow Matching (GFM) augments standard flow matching models with a diffusion term computed via a GNN, introducing an additional forward pass per ODE step. Each batch item corresponds to a node, yielding a graph with $B$ nodes per step. The computational overhead depends on the graph topology: fully connected graphs scale as $\mathcal{O}(B^2)$ due to dense attention, while $k$-nearest neighbor (KNN) graphs scale as $\mathcal{O}(Bk)$ in both compute and memory. The reaction and diffusion terms are parameterized by separate networks, with the diffusion module adding only $\sim 5$-$10\%$ to the total parameter count in our experiments. Importantly, GFM does not alter training objectives, solvers (e.g., Runge-Kutta, dopri5), or integration schedules. It preserves architectural compatibility while offering improved locality awareness at minimal additional cost.
\section{Experiments}
\label{sec:experiments}

\begin{table*}[t]
\footnotesize
\setlength{\tabcolsep}{5pt}
\centering
\begin{tabular}{llccccc}
\hline
\textbf{Dataset} & \textbf{Model} & \textbf{Total} & \boldmath$\mathbf{v}_{\mathit{diff}}$ & \textbf{FID} ($\downarrow$) & \textbf{FID} ($\downarrow$) &\textbf{Recall} \\
 & & \textbf{Parameters (M)} & \textbf{Parameters (M)} & \textbf{VAE$\rightarrow$Flow} & \textbf{True$\rightarrow$Flow} & \textbf{($\uparrow$)} \\
\hline
\multirow{6}{*}{\bfseries LSUN Church}
  & ADM (Baseline)   & 356.38 & 0     & --   & 7.70 & 0.39 \\
  & ADM+MPNN (Ours)  & 379.81 & 23.43 & 5.06 & 4.94 & 0.52 \\
  & ADM+GPS  (Ours)  & 374.20 & 18.13 & 4.67 & 4.61 & 0.51 \\
  \cline{2-7}
  & DiT  (Baseline)  & 456.80 & 0     & --   & 5.54 & 0.48 \\
  & DiT+MPNN (Ours)  & 502.09 & 45.29 & 2.90 & 2.92 & 0.65 \\
  & DiT+GPS  (Ours)  & 481.25 & 24.45 & 2.89 & 3.10 & 0.66 \\
\hline
\multirow{6}{*}{\bfseries FFHQ}
  & ADM (Baseline)   & 406.40 & 0     & --   & 8.07 & 0.40 \\
  & ADM+MPNN (Ours)  & 429.82 & 23.43 & 5.61 & 5.90 & 0.62 \\
  & ADM+GPS  (Ours)  & 424.53 & 18.13 & 4.80 & 5.09 & 0.62 \\
  \cline{2-7}
  & DiT  (Baseline)  & 456.80 & 0     & --   & 4.55 & 0.48 \\
  & DiT+MPNN (Ours)  & 480.23 & 23.43 & 3.80 & 4.52 & 0.66 \\
  & DiT+GPS  (Ours)  & 481.25 & 24.45 & 3.95 & 4.48 & 0.65 \\
\hline
\multirow{6}{*}{\bfseries LSUN Bedroom}
  & ADM (Baseline)   & 406.40 & 0     & --   & 7.05 & 0.39 \\
  & ADM+MPNN (Ours)  & 429.82 & 23.43 & 4.51 & 4.18 & 0.54 \\
  & ADM+GPS  (Ours)  & 424.53 & 18.13 & 4.26 & 3.97 & 0.56 \\
  \cline{2-7}
  & DiT  (Baseline)  & 456.80 & 0     & --   & 4.92 & 0.44 \\
  & DiT+MPNN (Ours)  & 480.23 & 23.43 & 2.55 & 2.66 & 0.64 \\
  & DiT+GPS  (Ours)  & 481.25 & 24.45 & 3.31 & 3.57 & 0.65 \\
\hline
\end{tabular}
\caption{Performance comparison against the baselines from \cite{dao2023flow} on LSUN Church, FFHQ, and LSUN Bedroom at $256\times256$ resolution. We report total parameters ($\mathbf{v}_{\text{react}} + \mathbf{v}_{\mathit{diff}}$), with $\mathbf{v}_{\mathit{diff}}$ parameters listed separately to show their minimal overhead, plus FID (lower is better) and recall (higher is better). Baseline parameter counts were computed from the \cite{dao2023flow} checkpoints on our machine. Corresponding \textit{FID(True$\rightarrow$VAE)} values: Church $=1.01$, FFHQ $=1.05$, Bedroom $=0.64$.}
\label{tab:model_performance_main}
\end{table*}

\begin{table*}[t]
\footnotesize
\setlength{\tabcolsep}{5pt}
\centering
\begin{tabular}{llccccc}
\hline
\textbf{Dataset} & \textbf{Model} & \textbf{Total} & \boldmath$\mathbf{v}_{\mathit{diff}}$ & \textbf{FID} ($\downarrow$) & \textbf{FID} ($\downarrow$) & \textbf{Recall} \\
 & & \textbf{Parameters (M)} & \textbf{Parameters (M)} & \textbf{VAE$\rightarrow$Flow} & \textbf{True$\rightarrow$Flow} & \textbf{($\uparrow$)} \\
\hline
\multirow{6}{*}{\bfseries AFHQ‐Cat}
  & ADM (Baseline)   & 153.10 & 0     & 7.40 & 7.34 & 0.32 \\
  & ADM+MPNN (Ours)  & 163.17 & 10.07 & 7.01 & 6.97 & 0.35 \\
  & ADM+GPS  (Ours)  & 171.23 & 18.13 & 2.92 & 4.63 & 0.54 \\
  \cline{2-7}
  & DiT (Baseline)   & 456.80 & 0     & 8.24 & 9.05 & 0.42 \\
  & DiT+MPNN (Ours)  & 480.23 & 23.43 & 7.19 & 8.07 & 0.51 \\
  & DiT+GPS  (Ours)  & 481.25 & 24.45 & 8.21 & 8.98 & 0.42 \\
\hline
\multirow{6}{*}{\bfseries CelebA‐HQ}
  & ADM (Baseline)   & 153.10 & 0     & 7.54 & 6.80 & 0.50 \\
  & ADM+MPNN (Ours)  & 176.53 & 23.43 & 5.33 & 6.28 & 0.56 \\
  & ADM+GPS  (Ours)  & 177.55 & 24.45 & 5.89 & 6.71 & 0.52 \\
  \cline{2-7}
  & DiT (Baseline)   & 456.80 & 0     & 6.85 & 6.23 & 0.55 \\
  & DiT+MPNN (Ours)  & 480.49 & 23.43 & 5.85 & 6.16 & 0.57 \\
  & DiT+GPS  (Ours)  & 481.25 & 24.45 & 6.09 & 6.18 & 0.59 \\
\hline
\end{tabular}
\caption{Performance comparison on AFHQ-Cat and CelebA-HQ at $256\times256$ resolution, computed entirely on our machine with $\mathbf{v}_{\text react}$ architectures ADM and DiT-L/2 being taken from \cite{dao2023flow}. We report total parameters ($\mathbf{v}_{\text{react}} + \mathbf{v}_{\mathit{diff}}$), with $\mathbf{v}_{\mathit{diff}}$ parameters listed separately to show their minimal overhead, plus FID (lower is better) and recall (higher is better). Corresponding \textit{FID(True$\rightarrow$VAE)} values for the datasets are: AFHQ-Cat $=3.18$, CelebA-HQ $=1.29$.}
\label{tab:model_performance_secondary}
\end{table*}

We evaluate Graph Flow Matching (GFM) on five standard unconditional image generation benchmarks: LSUN Church, LSUN Bedroom, FFHQ, AFHQ-Cat, and CelebA-HQ, all at $256 \times 256$ resolution. The goal is to assess the effect of incorporating neighbor-aware graph based correction into flow matching pipelines under consistent, state-of-the-art settings. We also conduct experiments with sparse K-nearest neighbor graphs (\Cref{sec:app_knn_ablation}), which demonstrate that GFM provides benefits even with sparse graphs.

\textbf{Experimental Settings.} To isolate the effect of the graph correction term, we retain all architecture and training settings from the latent flow matching (LFM) models of \citep{dao2023flow}. We therefore perform our experiments in the latent space of the pretrained VAE used by the aforementioned authors. Specifically, (i) {\bf Backbones}: The $\mathbf{v}_{\text react}(\mathbf{x}_{t},t)$ terms we use are the ADM U-Net~\citep{Dhariwal_ADM_NEURIPS2021_49ad23d1} and DiT Transformer~\citep{DiT_peebles2023scalable} (specifically, the ADM variants and DiT-L/2 variant from \cite{dao2023flow}), representing leading convolutional and attention-based flow architectures; (ii) {\bf Latent space}: The Stable Diffusion VAE~\citep{StableDiffusion_rombach2022high}, which maps $256 \times 256$ RGB images into $32 \times 32 \times 4$ latent tensors. All models operate exclusively in this latent space\footnote{The pretrained VAE used by \cite{dao2023flow}: https://huggingface.co/stabilityai/sd-vae-ft-mse}; (iii) {\bf Training strategy}: Constant-velocity flow matching loss (\Cref{eq:flow-loss}), the Dormand–Prince (dopri5) ODE integrator with $rtol=atol=10^{-5}$, and unmodified training hyperparameters from LFM \citep{dao2023flow}.

This ensures that any performance gain arises from our graph-based diffusion term (i.e., the graph correction module $\mathbf{v}_{\text{diff}}$) rather than hyperparameter tuning or architectural shifts. We evaluate the two implementations of $\mathbf{v}_{\text{diff}}$ discussed in \Cref{sec:method}. The {\it MPNN} implementation uses neural components $N_{\theta}^{(1)}$ and $N_{\theta}^{(2)}$, for which we employ identical U-Net \cite{UNet_ronneberger2015u} architectures derived from the U-Net design of \cite{PnPUNet_huang2021variational}. The {\it GPS} implementation, like MPNN, implements attention-based adjacency. Full architectural and hyperparameter details are provided in the Appendix.

    
To provide a nuanced assessment of generation quality, we introduce three FID metrics: (i) \textit{FID(True$\rightarrow$Flow)}, the standard FID computed between real and generated images; (ii) \textit{FID(True$\rightarrow$VAE)}, which measures the FID between real images and their reconstructions via the pretrained VAE (reported in table captions) to quantify reconstruction fidelity; and (iii) \textit{FID(VAE$\rightarrow$Flow)}, which compares VAE reconstructions with generated samples.
These metrics allow us to distinguish generative fidelity from the limitations of the VAE encoder-decoder pipeline and to better interpret results within the latent space modeling framework.

\Cref{tab:model_performance_main} and \Cref{tab:model_performance_secondary} report Fréchet Inception Distance (FID), recall, and parameter counts. Across all datasets and backbones, GFM consistently improves sample quality, achieving FID reductions of up to and sometimes exceeding $40\%$ relative to the base models. These improvements are achieved with only modest computational overhead—typically less than a $10\%$ increase in total parameters.

Qualitatively, GFM-enhanced models generate better-structured furniture and spatial layouts on LSUN Bedroom, sharper and more anatomically plausible faces on FFHQ, and more coherent architectural forms on LSUN Church as shown in Figures \ref{fig:ADM_FFHQ_comparisons_teaser} and Figure \ref{fig:lsun_bed_church_comparisons} (in the Appendix). The effectiveness of GFM across both convolutional (ADM) and transformer-based (DiT) models underscores its architectural generality. 

To further assess the role of graph structure, we conduct an ablation in which the adjacency matrix of the GPS module is replaced with the identity matrix, thereby eliminating inter-node communication while preserving architectural and parameter parity. Note that for the MPNN module, setting the adjacency matrix to identity would lead to a zero gradient matrix $\bfG$. This would effectively cancel out the diffusion term, leaving only the reaction term $\mathbf{v}_{\text{react}}$, making the MPNN architecture unsuitable for this particular ablation study. This ablation reduces GPS to a non-graph network and thus isolates the effect of the graph from the addition of a second network. Results in \Cref{tab:ablation_adjacency} show a clear degradation in FID and recall when graph connectivity is removed, confirming that {\bf performance gains stem from graph structure rather than parameter count or the addition of a second network alone.} This conclusion is further supported by our KNN ablation (\Cref{sec:app_knn_ablation}), where sparse graphs with controlled parameter counts still outperform wider baseline models, demonstrating that the benefits arise specifically from enabling inter-node communication, whether through dense or sparse connectivity, rather than simply from increased model capacity. Notably, GFM provides consistent performance gains across both convolutional (ADM) and transformer-based (DiT) architectures. This underscores the generality of GFM as an architectural enhancement that is agnostic to the flow matching backbone $\mathbf{v}_{\text{react}}$.


\begin{table*}[t]
\footnotesize
\setlength{\tabcolsep}{5pt}
\centering
\begin{tabular}{llccccc}
\hline
\textbf{Dataset} & \textbf{Model} & \textbf{Total} & \boldmath$\mathbf{v}_{\mathit{diff}}$ & \textbf{FID} ($\downarrow$) & \textbf{FID} ($\downarrow$) & \textbf{Recall} \\
 & & \textbf{Parameters (M)} & \textbf{Parameters (M)} & \textbf{VAE$\rightarrow$Flow} & \textbf{True$\rightarrow$Flow} & \textbf{($\uparrow$)} \\
\hline
\multirow{4}{*}{\bfseries LSUN Church}
  & ADM+GPS          & 374.20 & 18.13 & 4.67 & 4.61 & 0.51 \\
  & ADM+GPS (Adj=I)  & 374.20 & 18.13 & 6.24 & 5.71 & 0.51 \\
  \cline{2-7}
  & DiT+GPS          & 481.25 & 24.45 & 2.89 & 3.10 & 0.66 \\
  & DiT+GPS (Adj=I)  & 481.25 & 24.45 & 4.63 & 4.80 & 0.64 \\
\hline
\multirow{4}{*}{\bfseries FFHQ}
  & ADM+GPS          & 424.53 & 18.13 & 4.80 & 5.09 & 0.62 \\
  & ADM+GPS (Adj=I)  & 424.53 & 18.13 & 6.12 & 6.58 & 0.59 \\
  \cline{2-7}
  & DiT+GPS          & 481.25 & 24.45 & 3.95 & 4.48 & 0.65 \\
  & DiT+GPS (Adj=I)  & 481.25 & 24.45 & 4.02 & 4.53 & 0.64 \\
\hline
\multirow{4}{*}{\bfseries AFHQ‐Cat}
  & ADM+GPS          & 171.23 & 18.13 & 2.92 & 4.63 & 0.54 \\
  & ADM+GPS (Adj=I)  & 171.23 & 18.13 & 7.51 & 7.20 & 0.35 \\
  \cline{2-7}
  & DiT+GPS          & 481.25 & 24.45 & 8.21 & 8.98 & 0.42 \\
  & DiT+GPS (Adj=I)  & 481.25 & 24.45 & 8.26 & 9.01 & 0.41 \\
\hline
\end{tabular}
\caption{Ablation study isolating graph structure benefits in LSUN Church, FFHQ, and AFHQ-Cat. We compare our full graph augmented networks against an identical-parameter baseline where the graph adjacency matrix is set to identity (Adj=I), eliminating inter-node communication while preserving all network parameters. Total parameters include the $\mathbf{v}_{\mathit{diff}}$ parameters. Performance degradation when graph connectivity is removed (Adj=I) demonstrates that gains arise from graph structure, not just parameter count.}
\label{tab:ablation_adjacency}
\end{table*}

\section{Conclusion}

We introduced Graph Flow Matching (GFM), a lightweight architectural enhancement that improves the expressiveness of flow matching generative models by incorporating local neighborhood structure via a graph-based ``diffusion" term. Through a reaction–diffusion decomposition, GFM enables pointwise velocity predictors to aggregate contextual information from neighboring samples along the flow trajectory, offering a principled and scalable mechanism for improving the interpolation of the flow field.

Our results demonstrate that GFM consistently enhances generative quality across multiple datasets and architectures, reducing FID and increasing recall with minimal computational overhead. Importantly, these improvements hold across both convolutional and transformer-based backbones, highlighting the generality of our approach. 

Ablation studies confirm that performance gains arise from incorporating a graph, rather than merely from the addition of parameters or a second network. These findings underscore a broader principle: local correlations along the flow trajectory can be leveraged to improve sample quality.

Looking forward, we believe GFM opens several avenues for exploration and extensions to conditional or multimodal generation. More broadly, our work suggests that combining continuous-time generative frameworks with discrete geometric priors offers a promising direction for robust, high-fidelity generative modeling.

\bibliography{aaai2026}


\clearpage
\appendix

\section{Appendix}
\label{sec:appendix}

This appendix provides supplementary material including
further discussion, implementation details, and extended experimental results. We organize it into the following sections:
\begin{itemize}
    \item Section \ref{app:FromInterpToNeighborFlowPred}: Further discussion on the motivation for our framework.
    \item Section \ref{sec:experimental_details}: Experimental details
    \item Section \ref{sec:app_knn_ablation}: Diffusion-reaction with KNN ablation study
    \item Section \ref{sec:app_timing}: Image generation timing analysis
    \item Section \ref{sec:app_sota_comparisons}: Comparison against leading generative models.
    \item Section
    \ref{sec:app_QualitativeResults_SamplePictures}: Additional generated samples
\end{itemize}

\subsection{From Interpolation to Neighbor‑Aware Velocity Prediction}
\label{app:FromInterpToNeighborFlowPred}
The smoothness of the flow field $\mathbf{v}(\mathbf{x}, t)$ is a well established property in the literature. Under standard regularity conditions, such as a smooth Gaussian base distribution and a target distribution with finite moments, the learned flow field $\mathbf{v}(\mathbf{x},t)$ is typically Lipschitz continuous in both $\mathbf{x}$ and $t$, ensuring bounded gradients and stable interpolation trajectories~\cite{gao2024gaussian, gao2024convergence}. In theoretical analyses, sufficient regularity in $(\mathbf{x}, t)$ is often assumed to establish convergence guarantees~\cite{fukumizu2024flow}. Additionally, the learned vector field satisfies the continuity equation, ensuring mass-conserving evolution of the probability density over time~\cite{lipman2023flowmatchinggenerativemodeling, albergo2022building, gao2024gaussian}.

Viewed through this lens, flow matching may be interpreted as a scattered-data interpolation problem: the model observes velocities at a finite set of input points $(\mathbf{x}_{t_i}, t_i)$ and must reconstruct a smooth vector field over the entire domain. This perspective parallels classical interpolation methods in numerical analysis, such as radial basis function kernels~\cite{micchelli1984interpolation, wendland2004scattered}, splines~\cite{de1978practicalSplines}, or moving least squares~\cite{levin1998_MLSapproximation}, which all rely on smoothness and local support to stabilize predictions.

In these methods, incorporating local neighbor information, typically via compact support functions or interpolation stencils, improves approximation accuracy and generalization. For instance, radial basis function interpolation computes values at a query point as a weighted sum over nearby samples. Inspired by this, our neighbor-aware velocity module aggregates information from neighboring latent points via a graph-based message-passing mechanism, thereby instantiating a comparable inductive bias through a parameterized, data-driven mechanism.

Contemporary flow matching models often adopt a pointwise prediction scheme, directly regressing from $(\mathbf{x}_t, t)$ to $\mathbf{v}(\mathbf{x}_t, t)$ without modeling relationships across neighboring points. This resembles early point cloud models such as PointNet~\citep{qi2017pointnet}, which lacked mechanisms to capture local geometric structure, a limitation addressed by PointNet++~\citep{qi2017pointnetplusplus} through hierarchical neighborhood aggregation, and by DGCNN~\citep{DGCNN_paper}, which dynamically constructs graphs to model local relationships via edge convolutions. These architectures demonstrated the benefits of incorporating local context, an idea we adapt for velocity field prediction.

In high-dimensional settings, the absence of neighborhood modeling can hinder generalization, particularly in sparsely supervised, high curvature, or multimodal regions~\cite{Zhang_sparseSubspaceLearning, Folco_di2025highDLearning}. Given the smoothness of the velocity field $\mathbf{v}(\mathbf{x},t)$, points that are proximate in latent space may lie within the same semantic region and follow similar flow trajectories. Just as nearby points in classification tend to share labels, points close in $\mathcal{X} \times [0,1]$ may exhibit similar velocities. Leveraging this local correlation may help mitigate issues that arise in undersampled or complex regions.

\begin{figure*}[t]
  \centering
  \includegraphics[width=1.0\textwidth]{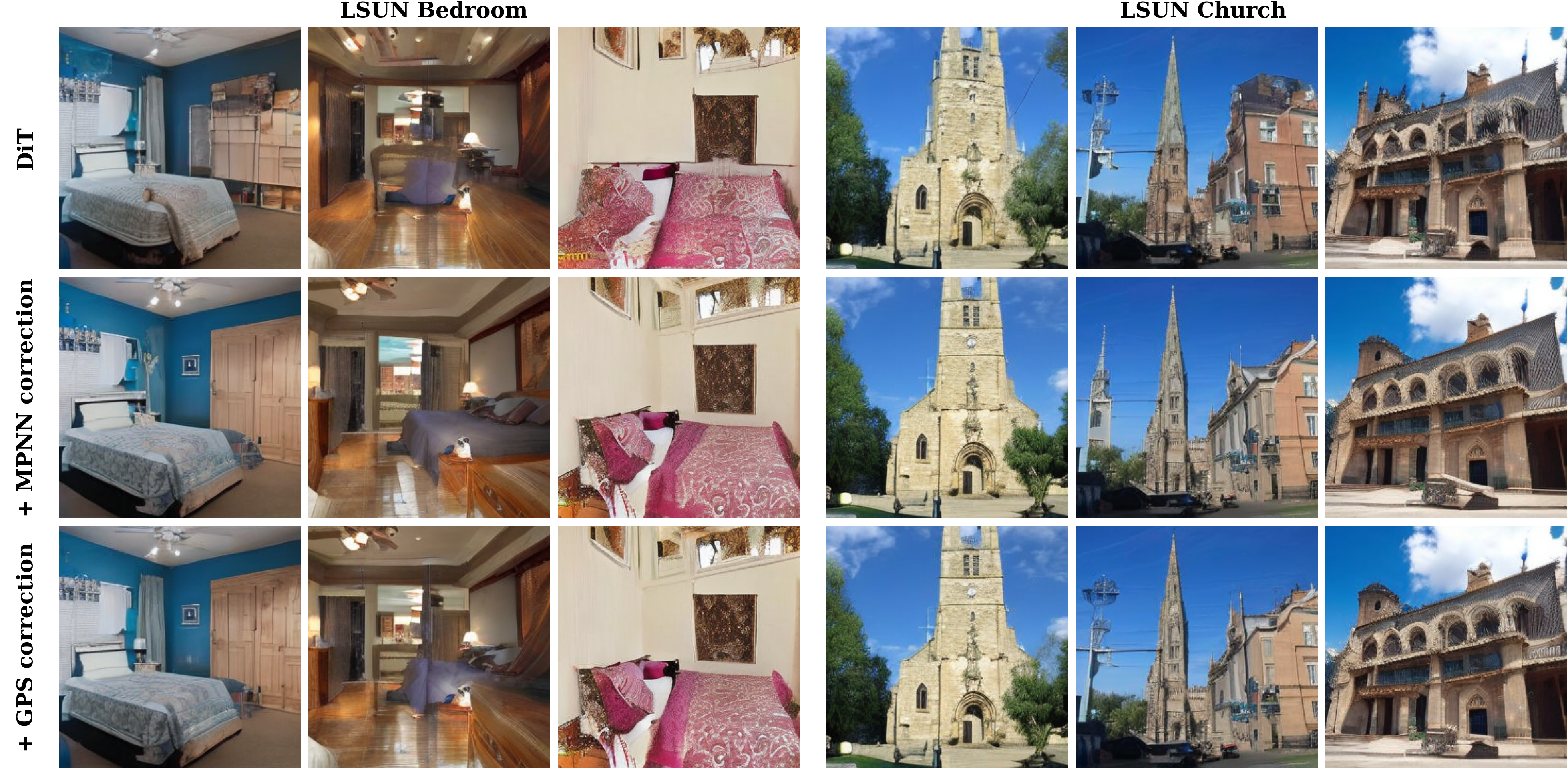}
    \caption{LSUN Bedroom and LSUN Church samples ($256\times256$) generated using the same random seed by: \textbf{(top)} baseline DiT-L/2 \cite{dao2023flow}, \textbf{(middle)} DiT-L/2 with MPNN-based correction, and \textbf{(bottom)} DiT-L/2 with GPS-based graph correction \cite{GPS_rampavsek2022recipe}. GFM variants generate more complete spatial structures and sharper boundaries compared to the baseline model.}
  \label{fig:lsun_bed_church_comparisons}
\end{figure*}

\section{Experimental Details}
\label{sec:experimental_details}
We implemented our code using PyTorch (offered under BSD-3 Clause license) and conducted experiments on Nvidia hardware (RTX 4090 for ADM UNet variants and RTX A6000 for DiT-L/2 experiments). All models from \cite{dao2023flow} were evaluated under identical conditions, with experiment tracking via Weights and Biases (wandb). The following subsections detail our datasets, architectures, and hyperparameter settings.

\subsection{Datasets}
We follow the exact preprocessing pipelines of \cite{dao2023flow} for all datasets, including VAE encoding at $256\times256$ resolution. To characterize the intrinsic variability of each dataset, we compute an internal FID score by randomly splitting each dataset into two non-overlapping halves and measuring the FID between them. Higher internal FID values indicate greater intrinsic variability within the dataset (\Cref{tab:appendix_internal_fid_scores}).

\begin{table}[!htbp]
\centering
\begin{tabular}{@{}lc@{}}
\toprule
\textbf{Dataset} & \textbf{Internal FID ($\downarrow$)} \\
\midrule
AFHQ-Cat      & 4.997 \\
CelebA-HQ     & 2.014 \\
FFHQ          & 1.123 \\
LSUN Church   & 0.416 \\
LSUN Bedroom & 0.020 \\
\bottomrule
\end{tabular}
\caption{Internal FID scores across datasets, computed as the FID between two non-overlapping halves of each dataset. Higher values indicate greater intrinsic variability.}
\label{tab:appendix_internal_fid_scores}
\end{table}

\paragraph{LSUN Church.}  
The LSUN Church dataset \cite{LSUN_dataset_yu15lsun} comprises approximately 126,227 training images of outdoor church buildings and architectural structures. This dataset presents particular challenges due to the complexity of architectural details and structural coherence required in the generated images.

\paragraph{LSUN Bedroom.}  
The LSUN Bedroom dataset \cite{LSUN_dataset_yu15lsun} contains around 3,033,042 training images of indoor bedroom scenes, making it substantially larger than the other datasets used in our experiments.  This dataset features complex indoor environments with various furniture arrangements, lighting conditions, and textures, requiring models to capture both local details and global room layouts.

\paragraph{FFHQ.}  
The Flickr-Faces-HQ (FFHQ) dataset \cite{FFHQ_dataset_karras2019style} consists of 70,000 high-quality human portrait photographs. This dataset is particularly challenging for generative models as human faces contain subtle details that are easily perceived when incorrectly generated, making it an excellent benchmark for evaluating generative fidelity.

\paragraph{AFHQ-Cat.}  
The AFHQ-Cat subset \cite{AFHQ_dataset_choi2020starganv2} contains $\sim$5,000 training images of cat faces from the Animal Faces-HQ dataset. Despite its smaller size compared to the other datasets, it presents unique challenges in capturing fine details of animal features and fur textures.

\paragraph{CelebA-HQ.}  
The CelebA-HQ dataset \cite{CelebAHQ_dataset_karras2018progressive} provides 30,000 celebrity face images. This dataset is a high-quality version of the original CelebA dataset, with improved resolution and reduced compression artifacts. CelebA-HQ features greater diversity in facial attributes, expressions, and backgrounds compared to FFHQ, though with a stronger bias toward frontal-facing poses and celebrity appearances.

For all datasets, we evaluate our models using standard metrics computed over 50,000 generated samples. The generative models operate in the latent space of the VAE, with final images obtained by decoding the generated latents.

\subsection{Training Algorithm}
We train our latent flow matching models as in LFM (\cite{dao2023flow}), where the authors used the \emph{reverse} convention that noise is at $t=1$ and the target distribution is at $t=0$, which we summarize here. At each iteration, a real image \(\mathbf{i}_0\) from the data distribution \(p_0\) is sampled and encoded into latent space via the VAE encoder \(E\), yielding \(\mathbf{x}_0 = E(\mathbf{i}_0)\). A sample from a standard Gaussian latent \(\mathbf{x}_1\sim\mathcal{N}(0,I)\) is then taken and a single time \(t\sim\mathcal{U}(0,1)\) is drawn.  The interpolation point \(\mathbf{x}_t = (1-t)\mathbf{x}_0 + t\mathbf{x}_1\) which lies along the straight‐line transport plan between \(\mathbf{x}_0\) and \(\mathbf{x}_1\) is constructed. Then, the instantaneous velocity \(\mathbf{v}_\theta(\mathbf{x}_t,t)\) is predicted with the flow network and the squared‐error loss against the ``true" transport velocity  \(\mathbf{x}_1 - \mathbf{x}_0\) is computed.  Finally, all network parameters are updated by gradient descent on this loss.

\subsection{Network configurations}

The network configurations for $\mathbf{v}_{\text{react}}$ for ADM from \cite{dao2023flow} are provided in \Cref{tab:adm_config} and that for DiT-L/2 can be found in \cite{dao2023flow} and in the corresponding GitHub repository.\footnote{\url{https://github.com/VinAIResearch/LFM/tree/main}} Those for $\mathbf{v}_{\text{diff}}$ are provided in \Cref{tab:app_mpnn_arch_hyperparams} and \Cref{tab:app_gps_arch_hyperparams}.

\subsubsection{MPNN Diffusion Architecture}
The MPNN diffusion module implements the velocity correction term $\mathbf{v}_{\text{diff}}$ using a message-passing neural network architecture inspired by the connection between graph neural networks and partial differential equations. As described in \Cref{eq:MPNN}, our MPNN architecture models the diffusion term as:

\begin{equation*}
\mathbf{v}_{\text{diff}}(\mathbf{x}_t, t) = -N_{\theta}^{(2)}\left( \mathbf{G}^\top\left[\sigma\left(\mathbf{G}\,N_{\theta}^{(1)}(\mathbf{x}_t, t)\right)\right],\, t \right),
\end{equation*}

where $\mathbf{G}$ is the graph gradient operator that computes differences between connected node features, $\sigma$ is a nonlinearity (we used the ELU function in all our experiments), and $N_{\theta}^{(1)}$ and $N_{\theta}^{(2)}$ are identical U-Net architectures (from \cite{PnPUNet_huang2021variational}, with code obtained from \cite{pnpFlow_martin2025pnpflow}) whose architectural hyperparameters are shown in Table \ref{tab:app_mpnn_arch_hyperparams}. The graph gradient $\mathbf{G}$ is constructed dynamically using attention-based adjacency, enabling the model to adapt its neighborhood structure based on the input latent representations. 

\begin{table*}[t]
\centering
\setlength{\tabcolsep}{3pt}
\begin{tabular}{@{}lccccc@{}}
\toprule
\textbf{Dataset}   & \textbf{$\mathbf{v}_{\text{react}}$} & \textbf{Base Channels} & \textbf{Channel Mult} & \textbf{Res Blocks} & \textbf{Attn.\ Resolutions} \\
\midrule
LSUN Church       & ADM & 85 & [1,2,2] & 2 & (32,16) \\
                   & DiT & 85 & [1,2,2] & 5 & (32,16) \\
\midrule
LSUN Bedroom      & ADM & 85 & [1,2,2] & 2 & (32,16) \\
                   & DiT & 85 & [1,2,2] & 2 & (32,16) \\
\midrule
FFHQ              & ADM & 85 & [1,2,2] & 2 & (32,16) \\
                   & DiT & 85 & [1,2,2] & 2 & (32,16) \\
\midrule
AFHQ‑Cat          & ADM & 70 & [1,2]   & 2 & (32,16) \\
                   & DiT & 85 & [1,2,2] & 2 & (32,16) \\
\midrule
CelebA‑HQ         & ADM & 85 & [1,2,2] & 2 & (32,16) \\
                   & DiT & 85 & [1,2,2] & 2 & (32,16) \\
\bottomrule
\end{tabular}
\caption{Architectural hyperparameters for models using MPNN as the diffusion term ($\mathbf{v}_{\text{diff}}$). Both $N_{\theta}^{(1)}$ and $N_{\theta}^{(2)}$ use identical U‑Net architectures from \cite{PnPUNet_huang2021variational} with the hyperparameters shown.}
\label{tab:app_mpnn_arch_hyperparams}
\end{table*}

\subsubsection{GPS Diffusion Architecture}
The GPS diffusion module implements the velocity correction term $\mathbf{v}_{\text{diff}}$ using a graph transformer architecture as described in \Cref{sec:method}. We adapted the implementation from the PyTorch Geometric repository\footnote{\url{https://github.com/pyg-team/pytorch_geometric/blob/master/examples/graph_gps.py}}. Table \ref{tab:app_gps_arch_hyperparams} presents the hyperparameters used for each dataset in our GPS module experiments. The module:

\begin{itemize}
  \item Projects latent tensors into a fixed-width embedding space (Hidden Dim)
  \item Processes node features through multiple graph transformer layers (Layers)
  \item Captures graph structure using random walk positional encodings of specified dimension and length (PE Dim, Walk Length)
  \item Applies temporal conditioning via learnable sinusoidal embeddings (Time Channels)
  \item Integrates information across nodes using multi-head attention mechanisms
\end{itemize}

The architecture combines GatedGraphConv operations for local message passing with transformer-style attention for global interactions. Graph connectivity is computed dynamically at each step through learnable attention weights, enabling adaptive neighborhood formation between latent representations.

\begin{table*}[t]
\footnotesize
\setlength{\tabcolsep}{3pt}
\centering
\begin{tabular}{lcccc}
\toprule
 & \textbf{FFHQ \& Bed (256)} & \textbf{Church (256)} & \textbf{CelebA-HQ$^{*}$} \\
\midrule
\# ResNet blocks per scale         & 2    & 2   & 2   \\
Base channels                      & 256  & 256 & 256 \\
Channel multiplier per scale       & 1,2,3,4 & 1,2,3,4 & 1,2,2,2 \\
Attention resolutions              & 16,8,4 & 16,8   & 16,8   \\
Channel multiplier for embeddings  & 4    & 4   & 4   \\
Label dimensions                   & 0    & 0   & 0   \\
\bottomrule
\end{tabular}
\caption{ADM \(\mathbf{v}_{\text{react}}\) configurations taken from \cite{dao2023flow}. The CelebA-HQ$^{*}$ configuration is the one obtained by testing the corresponding checkpoint from \cite{dao2023flow}.}
\label{tab:adm_config}
\end{table*}

\begin{table*}[t]
\small
\setlength{\tabcolsep}{3pt}
\centering
\begin{tabular}{lcccccc}
\toprule
\textbf{Dataset} & \textbf{$\mathbf{v}_{\text{react}}$} & \textbf{Hidden Dim} & \textbf{Layers} & \textbf{PE Dim} & \textbf{Walk Length} & \textbf{Time Channels} \\
\midrule
\multirow{2}{*}{LSUN Church} 
 & ADM & 256 & 16 & 16 & 20 & 64 \\
 & DiT & 300 & 16 & 16 & 20 & 64 \\
\midrule
\multirow{2}{*}{LSUN Bedroom} 
 & ADM & 256 & 16 & 16 & 20 & 64 \\
 & DiT & 300 & 16 & 16 & 20 & 64 \\
\midrule
\multirow{2}{*}{FFHQ} 
 & ADM & 256 & 16 & 16 & 20 & 64 \\
 & DiT & 300 & 16 & 16 & 20 & 64 \\
\midrule
\multirow{2}{*}{AFHQ-Cat} 
 & ADM & 256 & 16 & 16 & 20 & 64 \\
 & DiT & 300 & 16 & 16 & 20 & 64 \\
\midrule
\multirow{2}{*}{CelebA-HQ} 
 & ADM & 300 & 16 & 16 & 20 & 64 \\
 & DiT & 300 & 16 & 16 & 20 & 64 \\
\bottomrule
\end{tabular}
\caption{Architectural hyperparameters for GPS‐based diffusion term (\(\mathbf{v}_{\text{diff}}\)). All models used multihead attention with 4 heads and time embeddings with 8 learnable frequencies.}
\label{tab:app_gps_arch_hyperparams}
\end{table*}

\subsection{Training Hyperparameters}

We evaluated GFM using two reaction term architectures: DiT-L/2 and ADM U-Net. For both, we initialized $\mathbf{v}_{\text{react}}$ with corresponding architectures and checkpoints from \cite{dao2023flow}, then jointly trained the reaction and graph correction (``diffusion term") components. For AFHQ-Cat experiments, we used the CelebA-HQ checkpoints to initialize the reaction terms. 
For evaluation of performance metrics, FID scores were monitored every $3$K iterations, with training terminating after $80$K iterations without improvement. We saved models with the best FID scores and evaluated them using $50$K generated samples. 
For CelebA-HQ comparisons, we obtained higher, i.e., worse FID scores than reported in \cite{dao2023flow} when evaluating their provided checkpoints on our machines. To establish fair baselines, we trained these checkpoints for an additional $100$K iterations using the training settings in \cite{dao2023flow} and used the models with the lowest FID scores as the baseline models for the corresponding DiT and ADM experiments. 

Tables \ref{tab:app_dit_training_hyperparams} and \ref{tab:app_adm_training_hyperparams} detail the training hyperparameters for our GFM variants with DiT-L/2 and ADM U-Net backbones. All models used used the AdamW optimizer with $\beta_1=0.9$ and $\beta_2=0.999$. The original training settings for the \cite{dao2023flow} checkpoints are provided in Tables \ref{tab:vreact_adm_training_hparams} and \ref{tab:vreact_dit_training_hparams}. Similar to \cite{dao2023flow}, a cosine annealing scheduler was used for the learning rate.  
For ablation studies in Table \ref{tab:ablation_adjacency} with identity adjacency matrices, all other settings (besides their adjacency matrices) remained identical to their corresponding graph-based experiments.
\subsubsection{Seeds}
We set the initial seed during training to \(s_0 = 0\).  Whenever we generate images to compute FID at epoch \(e\), we reset the seed to \(s = s_0 + 10^6 + e\).  Finally, for the post-training evaluation on 50{,}000 generated samples, we fix the seed to \(s = 42\).

\begin{table*}[t]
\small
\centering
\setlength{\tabcolsep}{3.8pt}
\begin{tabular}{llccccc}
\toprule
$\mathbf{v}_{\text{diff}}$ & \textbf{Parameter}    & \textbf{LSUN Church} & \textbf{FFHQ} & \textbf{LSUN Bedroom} & \textbf{AFHQ‑Cat} & \textbf{CelebA‑HQ} \\
\midrule
\multirow{4}{*}{MPNN} 
 & Learning rate   & $10^{-4}$        & $2\cdot10^{-4}$ & $10^{-4}$        & $10^{-4}$        & $2\cdot10^{-4}$   \\
 & Batch size      & $42$             & $32$            & $32$             & $20$             & $32$              \\
 & Training iter.\ & $40$K            & $40$K           & $60$K            & $60$K            & $60$K             \\
 & \# GPUs         & $1$              & $1$             & $1$              & $1$              & $1$               \\
\midrule
\multirow{4}{*}{GPS}  
 & Learning rate   & $10^{-4}$        & $2\cdot10^{-4}$ & $10^{-4}$        & $10^{-4}$        & $2\cdot10^{-4}$   \\
 & Batch size      & $32$             & $32$            & $32$             & $32$             & $32$              \\
 & Training iter.\ & $40$K            & $40$K           & $40$K            & $60$K            & $60$K             \\
 & \# GPUs         & $1$              & $1$             & $1$              & $1$              & $1$               \\
\bottomrule
\end{tabular}
\caption{Training hyperparameters when \(\mathbf{v}_{\text{react}}\) is DiT‑L/2 across datasets. All models used the AdamW optimizer with \(\beta_1=0.9,\;\beta_2=0.999\).}
\label{tab:app_dit_training_hyperparams}
\end{table*}


\begin{table*}[t]
\small
\centering
\setlength{\tabcolsep}{3.8pt}
\begin{tabular}{llccccc}
\toprule
$\mathbf{v}_{\text{diff}}$ & \textbf{Parameter}    & \textbf{LSUN Church} & \textbf{FFHQ} & \textbf{LSUN Bedroom} & \textbf{AFHQ‑Cat} & \textbf{CelebA‑HQ} \\
\midrule
\multirow{4}{*}{MPNN} 
 & Learning rate   & $5\cdot10^{-5}$   & $2\cdot10^{-5}$   & $5\cdot10^{-5}$   & $5\cdot10^{-5}$   & $10^{-4}$        \\
 & Batch size      & $50$              & $128$             & $50$              & $85$              & $112$            \\
 & Training iter.\ & $50$K             & $120$K            & $40$K             & $200$K            & $60$K            \\
 & \# GPUs         & $1$               & $1$               & $1$               & $1$               & $1$              \\
\midrule
\multirow{4}{*}{GPS}  
 & Learning rate   & $5\cdot10^{-5}$   & $2\cdot10^{-5}$   & $5\cdot10^{-5}$   & $5\cdot10^{-5}$   & $10^{-4}$        \\
 & Batch size      & $128$             & $85$              & $85$              & $128$             & $112$            \\
 & Training iter.\ & $50$K             & $120$K            & $40$K             & $200$K            & $60$K            \\
 & \# GPUs         & $1$               & $1$               & $1$               & $1$               & $1$              \\
\bottomrule
\end{tabular}
\caption{Training hyperparameters when \(\mathbf{v}_{\text{react}}\) is ADM \cite{dao2023flow} across datasets. All models used the AdamW optimizer with \(\beta_1=0.9,\;\beta_2=0.999\).}
\label{tab:app_adm_training_hyperparams}
\end{table*}


\begin{table*}[t]
\small
\centering
\begin{tabular}{lccc}
\toprule
 & \textbf{CelebA 256} & \textbf{FFHQ} & \textbf{Church \& Bed} \\
\midrule
lr                                 & $5\cdot 10^{-5}$ & $2\cdot 10^{-5}$ & $5\cdot 10^{-5}$ \\
Adam optimizer ($\beta_1,\beta_2$) & $(0.9,0.999)$    & $(0.9,0.999)$    & $(0.9,0.999)$    \\
Batch size                         & $112$            & $128$            & $128$            \\
\# epochs                          & $600$            & $500$            & $500$            \\
\# GPUs                            & $2$              & $1$              & $1$              \\
\# training days                   & $1.3$            & $4.4$            & $6.9\;\&\;7.3$   \\
\bottomrule
\end{tabular}
\caption{Training hyperparameters for the ADM checkpoints used as \(\mathbf{v}_{\text{react}}\), reproduced exactly from \cite{dao2023flow}.}
\label{tab:vreact_adm_training_hparams}
\end{table*}


\begin{table*}[t]
\small
\centering
\begin{tabular}{lccc}
\toprule
 & \textbf{CelebA 256} & \textbf{FFHQ} & \textbf{Church \& Bed} \\
\midrule
Model                              & DiT-L/2        & DiT-L/2        & DiT-L/2         \\
lr                                 & $2\cdot10^{-4}$ & $2\cdot10^{-4}$ & $1\cdot10^{-4}$ \\
AdamW optimizer ($\beta_1,\beta_2$)& $(0.9,0.999)$  & $(0.9,0.999)$  & $(0.9,0.999)$   \\
Batch size                         & $32$           & $32$           & $96\;\&\;32$    \\
\# epochs                          & $500$          & $500$          & $500$           \\
\# GPUs                            & $1$            & $1$            & $2\;\&\;1$      \\
\# training days                   & $5.8$          & $5.5$          & $6.1\;\&\;12.5$ \\
\bottomrule
\end{tabular}
\caption{Training hyperparameters for the DiT checkpoints used as \(\mathbf{v}_{\text{react}}\), reproduced exactly from \cite{dao2023flow}.}
\label{tab:vreact_dit_training_hparams}
\end{table*}

\section{Ablation Study: Diffusion-Reaction with KNN}
\label{sec:app_knn_ablation}

In this ablation, we swap out our attention‐based adjacency for a simple $k$‐nearest‐neighbour (KNN) graph in latent space: for each node, we connect to its $k$ closest latents and assign each edge a learnable weight, so that in the diffusion term, information is aggregated strictly from a node's top‐$k$ neighbors instead of via attention scores. Concretely, we implement the diffusion term as a graph Laplacian multiplied by a per-channel diffusivity vector, mirroring the physical reaction–diffusion system in \Cref{eq:physical_reaction_diffusion_eq}. The overall velocity field remains the sum of the original flow-matching “reaction” network and this KNN-based “diffusion” operator. Let \(\mathbf{X}\in\mathbb{R}^{B\times C\times H\times W}\) be the batch of latent tensors at time \(t\). This network implements:

\begin{equation}
\label{eq:knn_ablation_network}
\begin{split}
\frac{d\mathbf{X}}{dt}(t) &= \mathbf{v}_{\mathrm{react}}\bigl(\mathbf{X}(t),t\bigr) \\[6pt]
&\quad+\;\underbrace{\kappa_{\theta}\bigl(\mathbf{X}(t),t\bigr)}_{\substack{\text{per‐channel}\\\text{diffusivity}}}
\;\odot\;\bigl[\mathbf{L}(\mathbf{X}(t))\,\mathbf{X}_{\mathrm{flat}}(t)\bigr]\,
\end{split}
\end{equation}

where:
\begin{itemize}[leftmargin=1.5em]
  \item \(\mathbf{v}_{\mathrm{react}}(\mathbf{X},t)\) is a UNet variant from \cite{PnPUNet_huang2021variational}, with code taken from the GitHub repository of \cite{pnpFlow_martin2025pnpflow}.
  \item \(\kappa_{\theta}(\mathbf{X},t)\in\mathbb{R}^{B\times C}\) is a \emph{per‐example, per‐channel} diffusivity vector produced by
    \[
      \mathbf{Z} = \mathrm{UNet}(\mathbf{X},t),\quad
      \kappa_{\theta}(\mathbf{X},t) = \sigma\bigl(\mathrm{GAP}(\mathbf{Z})\bigr)\,,
    \]
    where \(\mathrm{GAP}\) is global average pooling to shape \((B,C)\) and \(\sigma\) is a smooth activation (we used ELU in our experiments).
  \item \(\mathbf{L}(\mathbf{X})=\mathbf{I} - \mathbf{D}^{-1/2}\mathbf{A}\,\mathbf{D}^{-1/2}\) is the \emph{normalized graph Laplacian} on the batch.  We build the adjacency \(\mathbf{A}\in\mathbb{R}^{B\times B}\) by:
    \begin{enumerate}[leftmargin=2em]
      \item Computing pairwise cosine similarities \(\mathrm{sim}_{ij} = \frac{\mathbf{X}_i\cdot \mathbf{X}_j}{\|\mathbf{X}_i\|\|\mathbf{X}_j\|}\).
      \item For each node \(i\), selecting its top-\(K\) neighbors and assigning each neighbor rank \(r\in\{0,\dots,K\}\) a learnable weight \(w_r\).
      \item Masking to retain only these \(K\) edges, then row-normalizing so that each row of \(\mathbf{A}\) sums to one.
    \end{enumerate}
  \item \(\mathbf{X}_{\mathrm{flat}}\in\mathbb{R}^{B\times (C\,H\,W)}\) denotes each tensor flattened along spatial dimensions, and “\(\odot\)” broadcasts \(\kappa_\theta\) across spatial axes.
\end{itemize}

\Cref{eq:knn_ablation_network} thus decomposes the velocity into a standard flow‐matching “reaction” term $\mathbf{v}_{\mathrm{react}}$ and a KNN‐based “diffusion” term given by the graph Laplacian $\mathbf{L}(\mathbf{X})$ scaled per‐channel by $\kappa_\theta(\mathbf{X},t)$. This KNN diffusion introduces local coupling across the batch at a cost of $\mathcal{O}(B\,K)$ per step.
 
To isolate the impact of graph structure from parameter count, we conducted a controlled ablation study on AFHQ-Cat using KNN-based graph diffusion. The architectural hyperparameters for the networks used are provided in \Cref{tab:app_knn_ablation_hyperparams}. All models were trained for up to 2500 epochs with early stopping, using a patience of 400 epochs (training was terminated if the validation FID did not improve for 400 consecutive epochs) and a learning rate of $10^{-4}$. \Cref{tab:app_ablation_model_performance_knn} presents these results, comparing GFM variants (with $k=5,10,20$ neighbors) against baseline models without diffusion terms ("noDiff" in the table). Notably, we ensured the baseline models had slightly more parameters ($18.62$M vs. $18.47$M) by increasing their width (number of channels), eliminating the possibility that improved performance stems merely from additional capacity. KNN-based GFM variants consistently outperform these wider baselines across all metrics, demonstrating that neighborhood structure rather than parameter count drives improvement.


\begin{table*}[t]
\centering
\footnotesize
\setlength{\tabcolsep}{3.5pt}
\begin{tabular}{lccccccc}
\toprule
\textbf{Model} & \textbf{Batch Size} & \textbf{Total Parameters (M)} & \textbf{$\mathbf{v}_{\text{diff}}$ (M)} & \textbf{FID (VAE$\rightarrow$Flow $\downarrow$)} & \textbf{FID (True$\rightarrow$Flow $\downarrow$)} & \textbf{KID ($\downarrow$)} & \textbf{Recall ($\uparrow$)} \\
\midrule
\textbf{noDiff} (seed = 0)   & 100 & 18.62 & 0    & 16.96 & 17.78 & $1.37\times10^{-2}$ & 0.21 \\
\textbf{noDiff} (seed = 100) &  64 & 18.62 & 0    & 15.72 & 15.26 & $1.17\times10^{-2}$ & 0.32 \\
\midrule
\textbf{KNN ($k=5$)}  &  85 & 18.47 & 5.54 & 12.90 & 12.32 & $8.07\times10^{-3}$ & 0.38 \\
\textbf{KNN ($k=10$)} &  64 & 18.47 & 5.54 & 11.82 & 11.33 & $7.65\times10^{-3}$ & 0.38 \\
\textbf{KNN ($k=20$)} &  64 & 18.47 & 5.54 & 11.64 & 10.91 & $6.94\times10^{-3}$ & 0.36 \\
\bottomrule
\end{tabular}
\caption{Ablation study comparing KNN‑based GFM variants against baseline models that use only the reaction term $\mathbf{v}_{\text{react}}$ on AFHQ‑Cat at $256\times256$ resolution. Two baseline models (“noDiff”) are shown with different random seeds (0 and 100) to account for initialization variance. The baseline models contain 18.62M parameters, while GFM variants have 18.47M parameters. Samples were generated using Runge‑Kutta (RK4) integration with 3 steps. Evaluation metrics are computed on 50,000 generated samples. Corresponding \emph{FID(True$\rightarrow$VAE)} for AFHQ‑Cat = 3.18.}
\label{tab:app_ablation_model_performance_knn}
\end{table*}


\begin{table*}[t]
\centering
\small
\setlength{\tabcolsep}{6pt}
\begin{tabular}{lccccc}
\toprule
\textbf{Model} & \textbf{Component} & \textbf{Base Channels} & \textbf{Channel Mult} & \textbf{Res Blocks} & \textbf{Attn. Resolutions} \\
\midrule
\multirow{2}{*}{\textbf{KNN} ($k=5,10,20$)}
  & $\mathbf{v}_{\text{react}}$                      & 100  & [1,2] & 3 & (32,16) \\
  & $\kappa_{\theta}$-net ($\mathbf{v}_{\text{diff}}$) &  54  & [1,2] & 5 & (32,16) \\
\midrule
\textbf{noDiff}
  & $\mathbf{v}_{\text{react}}$ only                & 120  & [1,2] & 3 & (32,16) \\
\bottomrule
\end{tabular}
\caption{Architectural hyperparameters for the KNN-based diffusion ablation study (see \Cref{sec:app_knn_ablation}); results in \Cref{tab:app_ablation_model_performance_knn}. Models use the U-Net of \cite{PnPUNet_huang2021variational} and were trained for 2,500 epochs with early stopping (patience = 400).}
\label{tab:app_knn_ablation_hyperparams}
\end{table*}


\section{Timing of Image Generation}
\label{sec:app_timing}
The ADM runs were conducted on an Nvidia RTX 4090 (24 GB) GPU and the DiT runs on an Nvidia A6000 (48 GB) GPU. We measured the time required for image generation in \Cref{tab:app_sampling_costs} using the Nvidia A6000 GPU for all models. The number of function evaluations (NFE) and times required to generate 1 image are reported. The corresponding timing results for the checkpoints from \cite{dao2023flow} were evaluated on our GPU.

As shown in \Cref{tab:app_sampling_costs}, GFM models require approximately the same number of function evaluations as the baselines, with most variants showing statistically indistinguishable NFEs (confidence intervals overlap) or even slightly lower NFEs in some variants. The primary computational overhead comes from the additional graph processing, which increases sampling time. While there is a modest increase in wall-clock time, this also yields substantially improved performance as can be seen in the table. For example, DiT+MPNN achieves a 47\% reduction in FID (from 5.54 to 2.92) on LSUN Church while maintaining comparable NFE, demonstrating that graph-based corrections significantly boost generation quality without additional solver steps and with minimal extra parameters (\Cref{tab:model_performance_main}, \Cref{tab:model_performance_secondary}).

\begin{table*}[t]
  \centering
  \footnotesize
  \setlength{\tabcolsep}{5pt}
  \begin{tabular}{@{}l l r@{$\,\pm\,$}l r@{$\,\pm\,$}l r r@{}}
    \toprule
    \textbf{Dataset} & \textbf{Model} & \multicolumn{2}{c}{\textbf{NFE } ($\downarrow$)} & \multicolumn{2}{c}{\textbf{Time (s)} ($\downarrow$)} & \textbf{FID} ($\downarrow$) & \textbf{Recall} ($\uparrow$) \\
    \midrule
    \multirow{6}{*}{LSUN Church} 
      & ADM                       & $83.00$  & $7.71$   & $0.61$  & $0.55$   & $7.70$ & $0.39$ \\
      & ADM+MPNN$^{\dagger}$      & $89.00$  & $15.24$  & $1.63$  & $0.28$   & $4.94$ & $0.52$ \\
      & ADM+GPS$^{\dagger}$       & $82.40$  & $13.73$  & $1.38$  & $0.23$   & $4.61$ & $0.51$ \\
      & DiT                       & $87.80$  & $10.75$  & $1.34$  & $0.16$   & $5.54$ & $0.48$ \\
      & DiT+MPNN$^{\dagger}$      & $100.40$ & $14.25$  & $4.94$  & $0.70$   & $2.92$ & $0.65$ \\
      & DiT+GPS$^{\dagger}$       & $93.80$  & $10.75$  & $2.81$  & $0.32$   & $3.10$ & $0.66$ \\
    \addlinespace
    \midrule
    \midrule
    \multirow{6}{*}{LSUN Bedroom} 
      & ADM                       & $91.40$  & $15.05$  & $0.93$  & $0.15$   & $7.05$ & $0.39$ \\
      & ADM+MPNN$^{\dagger}$      & $83.00$  & $14.01$  & $1.76$  & $0.29$   & $4.18$ & $0.54$ \\
      & ADM+GPS$^{\dagger}$       & $77.00$  & $10.82$  & $1.51$  & $0.21$   & $3.97$ & $0.56$ \\
      & DiT                       & $90.20$  & $20.09$  & $1.38$  & $0.30$   & $4.92$ & $0.44$ \\
      & DiT+MPNN$^{\dagger}$      & $95.60$  & $18.43$  & $2.98$  & $0.57$   & $2.66$ & $0.64$ \\
      & DiT+GPS$^{\dagger}$       & $98.60$  & $24.07$  & $2.96$  & $0.71$   & $3.57$ & $0.65$ \\
    \addlinespace
    \midrule
    \midrule
    \multirow{6}{*}{FFHQ} 
      & ADM                       & $83.60$  & $6.68$   & $0.85$  & $0.07$   & $8.07$ & $0.40$ \\
      & ADM+MPNN$^{\dagger}$      & $69.20$  & $8.82$   & $1.34$  & $0.17$   & $5.90$ & $0.62$ \\
      & ADM+GPS$^{\dagger}$       & $79.40$  & $10.55$  & $1.55$  & $0.20$   & $5.09$ & $0.62$ \\
      & DiT                       & $89.00$  & $8.16$   & $1.36$  & $0.12$   & $4.55$ & $0.48$ \\
      & DiT+MPNN$^{\dagger}$      & $91.40$  & $10.88$  & $2.82$  & $0.33$   & $4.52$ & $0.66$ \\
      & DiT+GPS$^{\dagger}$       & $87.80$  & $11.40$  & $2.62$  & $0.34$   & $4.48$ & $0.65$ \\
    \addlinespace
    \midrule
    \midrule
    \multirow{6}{*}{AFHQ-Cat} 
      & ADM                       & $81.80$  & $8.07$   & $0.60$  & $0.06$   & $7.34$ & $0.32$ \\
      & ADM+MPNN$^{\dagger}$      & $83.60$  & $5.50$   & $1.51$  & $0.10$   & $6.97$ & $0.35$ \\
      & ADM+GPS$^{\dagger}$       & $126.80$ & $13.89$  & $2.15$  & $0.23$   & $4.63$ & $0.54$ \\
      & DiT                       & $77.60$  & $5.50$   & $1.18$  & $0.08$   & $9.05$ & $0.42$ \\
      & DiT+MPNN$^{\dagger}$      & $78.20$  & $8.92$   & $2.45$  & $0.28$   & $8.07$ & $0.51$ \\
      & DiT+GPS$^{\dagger}$       & $84.20$  & $11.71$  & $2.50$  & $0.34$   & $8.98$ & $0.42$ \\
    \addlinespace
    \midrule
    \midrule
    \multirow{6}{*}{CelebA-HQ} 
      & ADM                       & $84.20$  & $10.75$  & $0.61$  & $0.08$   & $6.80$ & $0.50$ \\
      & ADM+MPNN$^{\dagger}$      & $82.40$  & $9.37$   & $1.46$  & $0.16$   & $6.28$ & $0.56$ \\
      & ADM+GPS$^{\dagger}$       & $78.80$  & $5.88$   & $1.34$  & $0.10$   & $6.71$ & $0.52$ \\
      & DiT                       & $91.40$  & $12.13$  & $1.39$  & $0.18$   & $6.23$ & $0.55$ \\
      & DiT+MPNN$^{\dagger}$      & $85.40$  & $7.80$   & $2.62$  & $0.24$   & $6.16$ & $0.57$ \\
      & DiT+GPS$^{\dagger}$       & $81.20$  & $7.00$   & $2.44$  & $0.21$   & $6.18$ & $0.59$ \\
    \bottomrule
  \end{tabular}
\caption{Sampling cost in terms of number of function evaluations (NFE), average sampling time per sample (in seconds), FID (True$\rightarrow$Flow) scores, and recall values for each model on five benchmarks at $256\times256$ resolution. NFE and timing statistics are computed over 10 independent runs. Models marked with $\dagger$ denote baseline architectures augmented with a GNN-based diffusion term; they incur modest computational overhead but consistently achieve substantially better image quality (lower FID) and diversity (higher recall) across all datasets.}
\label{tab:app_sampling_costs}
\end{table*}

\section{Comparison with Leading Generative Models}
\label{sec:app_sota_comparisons}

\Cref{tab:appendix_comparisons_with_sota} compares our GFM variants—ADM+MPNN, ADM+GPS, DiT+MPNN, and DiT+GPS—to a range of leading generative models, including deterministic flow matching (FM)~\citep{lipman2023flowmatchinggenerativemodeling}, latent diffusion (LDM)~\citep{StableDiffusion_rombach2022high}, wavelet diffusion (WaveDiff)~\citep{WaveDiff_phung2023wavelet}, DDPM~\citep{DDPM_ho2020denoising}, ImageBART~\citep{ImageBART_esser2021imagebart}, and top GANs (StyleGAN~\citep{FFHQ_dataset_karras2019style}, StyleGAN2~\citep{StyleGAN2_karras2020training}, ProjectedGAN~\citep{ProjectedGAN_sauer2021projected}, DiffGAN~\citep{DiffGAN_wangdiffusion}). As can be observed, augmenting ADM and DiT backbones with our lightweight graph correction term consistently improves FID and recall over their baseline pointwise counterparts. Notably, these ADM+GFM and DiT+GFM models achieve FID and recall on par with the diffusion and GAN pipelines in \Cref{tab:appendix_comparisons_with_sota}, showing that simply adding GFM as a graph based correction yields significant quality gains, without altering the core architecture or training objective.

\begin{table*}[t]
\footnotesize
\centering
\setlength{\tabcolsep}{4pt}
\begin{tabular}{@{}lccccc@{}}
\toprule
Dataset/Model & Total Parameters & $\mathbf{v}_{\text{diff}}$ & FID ($\downarrow$) & FID ($\downarrow$) & Recall \\
 & \textbf{(M)} & \textbf{(M)} & \textbf{VAE$\rightarrow$Flow} & \textbf{True$\rightarrow$Flow} & \textbf{($\uparrow$)} \\
\midrule
\multicolumn{6}{@{}l}{\textbf{LSUN Church}} \\
ADM~\cite{dao2023flow} (Baseline) & $356.38$ & $0$ & - & $7.70$ & $0.39$ \\
ADM+MPNN(Ours) & $379.81$ & $23.43$ & $5.06$ & $4.94$ & $0.52$ \\
ADM+GPS (Ours)& $374.20$ & $18.13$ & $4.67$ & $4.61$ & $0.51$ \\
\midrule
DiT~\cite{dao2023flow} (Baseline)& $456.80$ & $0$ & - & $5.54$ & $0.48$ \\
DiT+MPNN (Ours)& $502.09$ & $45.29$ & $2.90$ & $2.92$ & $0.65$ \\
DiT+GPS (Ours)& $481.25$ & $24.45$ & $2.89$ & $3.10$ & $0.66$ \\
\midrule
FM \cite{lipman2023flowmatchinggenerativemodeling} & - & $0$ & - & $10.54$ & - \\
LDM \cite{StableDiffusion_rombach2022high}& - & $0$ & - & $4.02$ & $0.52$ \\
WaveDiff \cite{WaveDiff_phung2023wavelet} & - & $0$ & - & $5.06$ & $0.40$ \\
DDPM \cite{DDPM_ho2020denoising} & - & 0 & - & $7.89$ & - \\
ImageBART \cite{ImageBART_esser2021imagebart} & - & $0$ & - & $7.32$ & - \\
StyleGAN \cite{FFHQ_dataset_karras2019style}& - & $0$ & - & $4.21$ & - \\
StyleGAN2 \cite{StyleGAN2_karras2020training} & - & $0$ & - & $3.86$ & $0.36$ \\
ProjectedGAN \cite{ProjectedGAN_sauer2021projected} & - & $0$ & - & $1.59$ & $0.44$ \\
\midrule
\midrule
\multicolumn{6}{@{}l}{\textbf{FFHQ}} \\
ADM~\cite{dao2023flow} (Baseline)& $406.40$ & $0$ & - & $8.07$ & $0.40$ \\
ADM+MPNN (Ours) & $429.82$ & $23.43$ & $5.61$ & $5.90$ & $0.62$ \\
ADM+GPS (Ours)& $424.53$ & $18.13$ & $4.80$ & $5.09$ & $0.62$ \\
\midrule
DiT~\cite{dao2023flow} (Baseline)& $456.80$ & $0$ & -  & $4.55$ & $0.48$ \\
DiT+MPNN (Ours)& $480.23$ & $23.43$ & $3.80$ & $4.52$ & $0.66$ \\
DiT+GPS (Ours)& $481.25$ & $24.45$ & $3.95$ & $4.48$ & $0.65$ \\
\midrule
LDM \cite{StableDiffusion_rombach2022high}& - & $0$ & - & $4.98$ & $0.50$ \\
ImageBART \cite{ImageBART_esser2021imagebart} & - & $0$ & - & $9.57$ & $-$ \\
ProjectedGAN \cite{ProjectedGAN_sauer2021projected}& - & $0$ & - & $3.08$ & $0.46$ \\
StyleGAN \cite{FFHQ_dataset_karras2019style} & - & $0$ & - & $4.16$ & $0.46$ \\
\midrule
\midrule
\multicolumn{6}{@{}l}{\textbf{LSUN Bedroom}} \\
ADM~\cite{dao2023flow} (Baseline)& $406.40$ & $0$ & - & $7.05$ & $0.39$ \\
ADM+MPNN (Ours)& $429.82$ & $23.43$ & $4.51$ & $4.18$ & $0.54$ \\
ADM+GPS (Ours)& $424.53$ & $18.13$ & $4.26$ & $3.97$ & $0.56$ \\
\midrule
DiT~\cite{dao2023flow} (Baseline)& $456.80$ & $0$ & - & $4.92$ & $0.44$ \\
DiT+MPNN (Ours)& $480.23$ & $23.43$ & $2.55$ & $2.66$ & $0.64$ \\
DiT+GPS (Ours)& $481.25$ & $24.45$ & $3.31$ & $3.57$ & $0.65$ \\
\midrule
LDM \cite{StableDiffusion_rombach2022high}& - & $0$ & - & $2.95$ & $0.48$ \\
DDPM \cite{DDPM_ho2020denoising}& - & $0$ & - & $4.90$ & $-$ \\
ImageBART \cite{ImageBART_esser2021imagebart} & - & $0$ & - & $5.51$ & $-$ \\
ADM \cite{Dhariwal_ADM_NEURIPS2021_49ad23d1} & - & $0$ & - & $1.90$ & $0.51$ \\
PGGAN \cite{PGGAN_karras2018progressive} & - & $0$ & - & $8.34$ & $-$ \\
StyleGAN \cite{FFHQ_dataset_karras2019style}& - & $0$ & - & $2.35$ & $0.48$ \\
ProjectedGAN\cite{ProjectedGAN_sauer2021projected} & - & $0$ & - & $1.52$ & $0.34$ \\
DiffGAN \cite{DiffGAN_wangdiffusion}& - & $0$ & - & $1.43$ & $0.58$ \\
\bottomrule
\end{tabular}
\caption{Performance comparison against a range of leading generative models and the corresponding $\mathbf{v}_\text{react}$-only baselines on LSUN Church, FFHQ, and LSUN Bedroom at $256\times256$ resolution. We report total parameters, diffusion‐term ($\mathbf{v}_{\text{diff}}$‐net) parameters, FID (lower is better), and recall (higher is better). Baseline parameter counts were measured from the \cite{dao2023flow} checkpoints on our hardware; our results are highlighted in light blue. Corresponding {\it FID(True$\rightarrow$VAE)} values are: LSUN Church $=1.01$, FFHQ $=1.05$, LSUN Bedroom $=0.64$.}
\label{tab:appendix_comparisons_with_sota}
\end{table*}

\section{Qualitative Results}
\label{sec:app_QualitativeResults_SamplePictures}

\begin{figure}[ht]
  \centering
  \includegraphics[width=\linewidth]{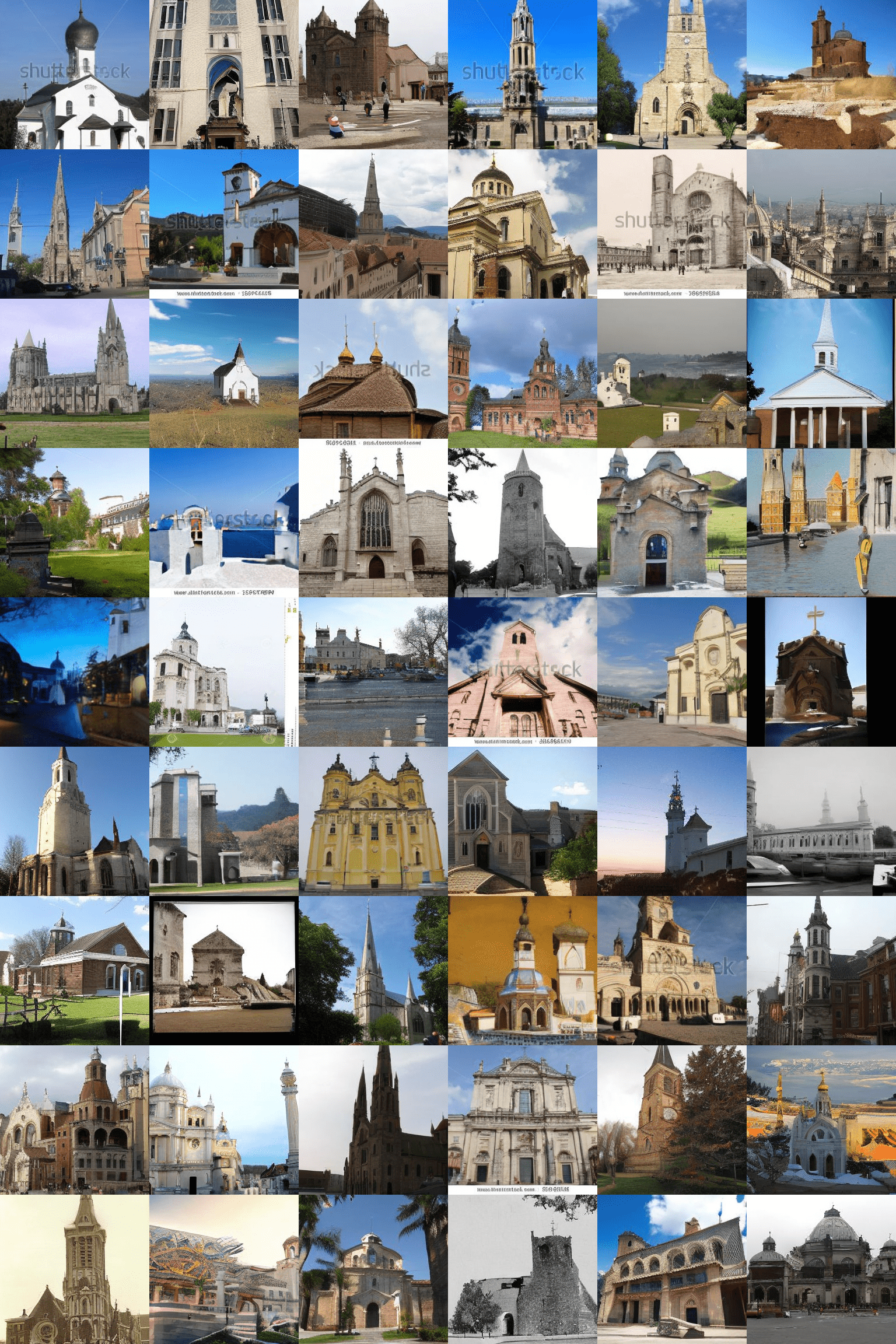}
  \caption{Randomly generated samples of LSUN Church using DiT-L/2+MPNN}
  \label{fig:app_LSUN_Church_samples}
\end{figure}

\begin{figure}[ht]
  \centering
  \includegraphics[width=\linewidth]{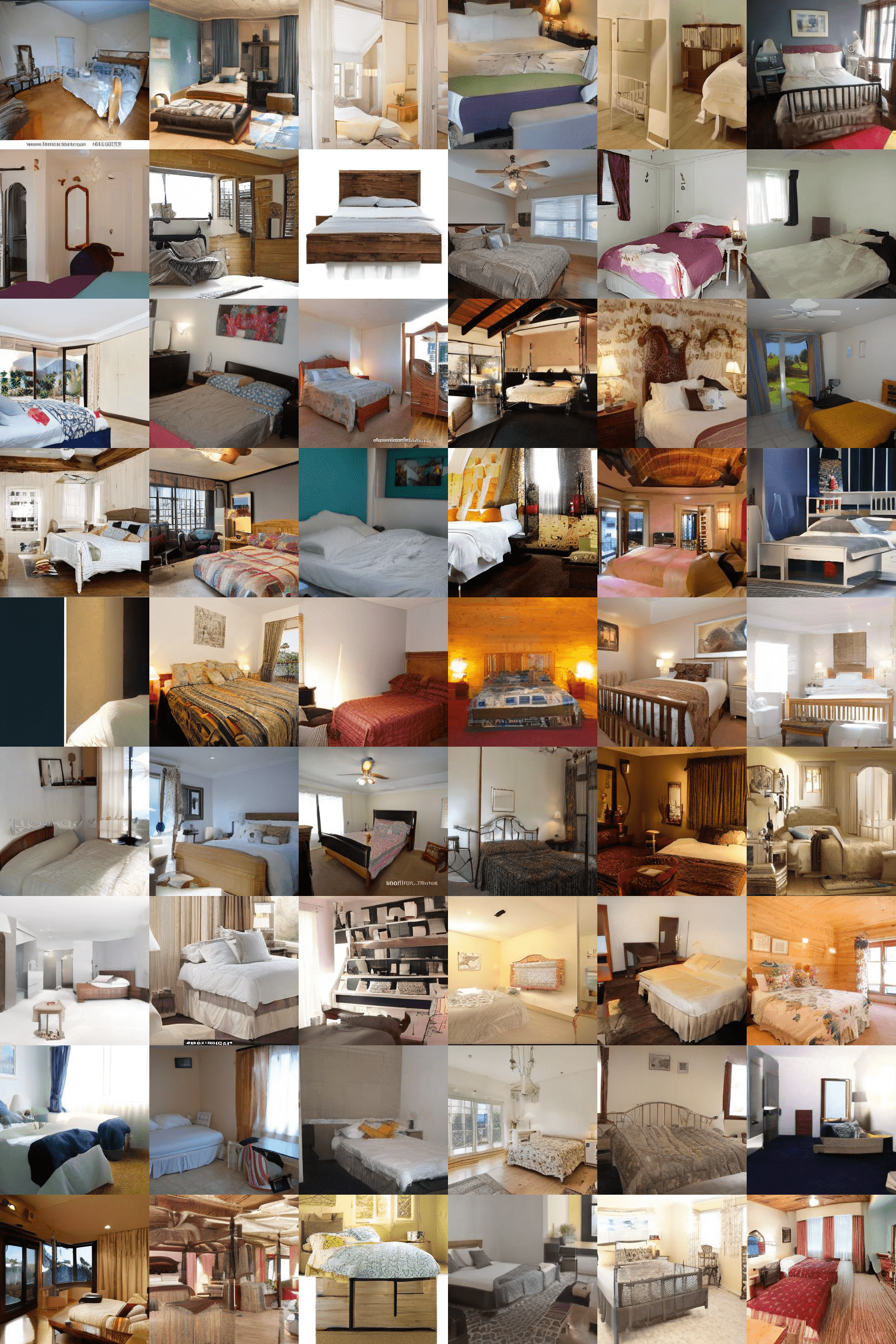}
  \caption{Randomly generated samples of LSUN Bedroom using DiT-L/2+MPNN}
  \label{fig:app_LSUN_Bedroom_samples}
\end{figure}

\begin{figure}[ht]
  \centering
  \includegraphics[width=\linewidth]{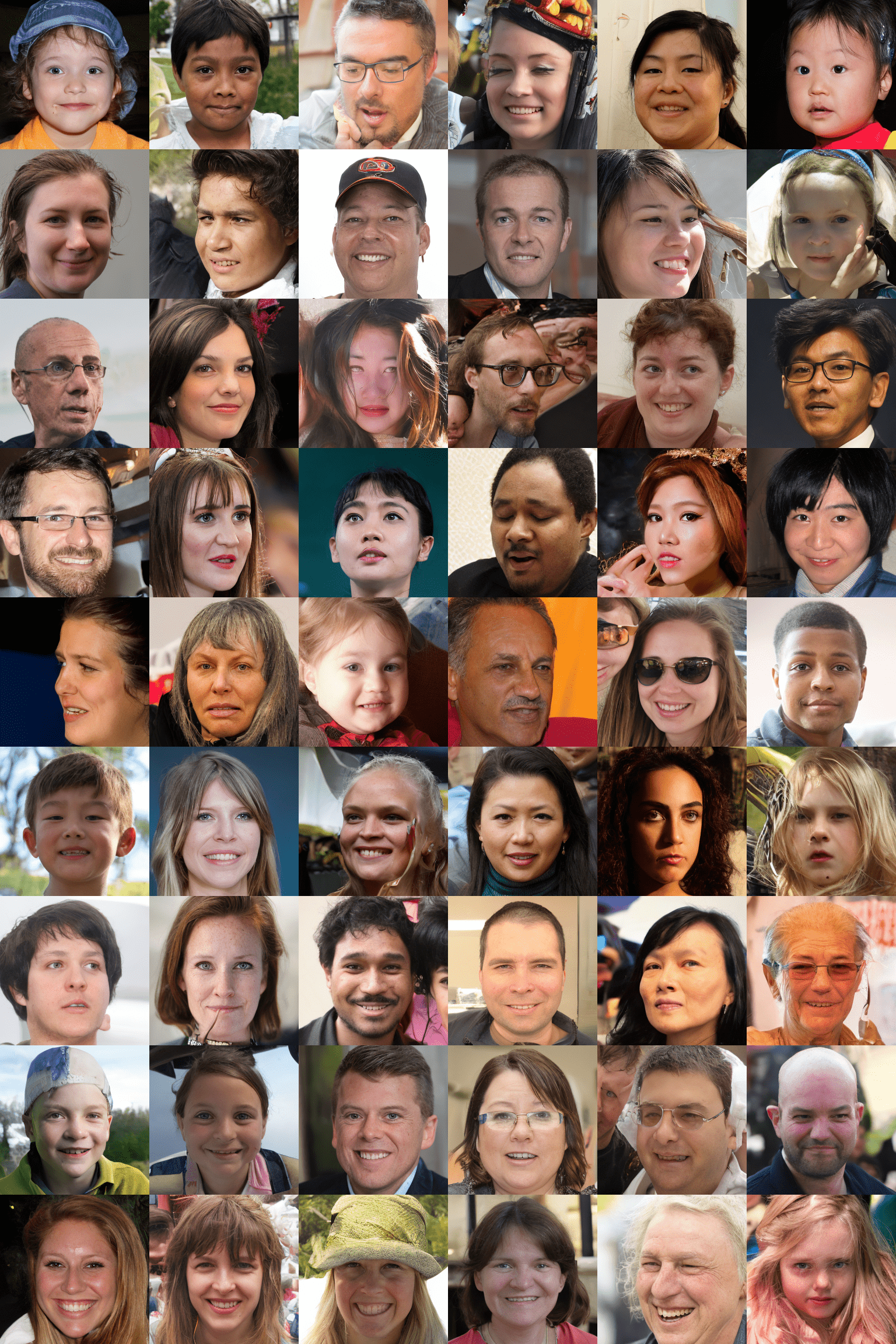}
  \caption{Randomly generated samples of FFHQ using DiT-L/2+GPS}
  \label{fig:app_FFHQ_samples}
\end{figure}

\begin{figure}[ht]
  \centering
  \includegraphics[width=\linewidth]{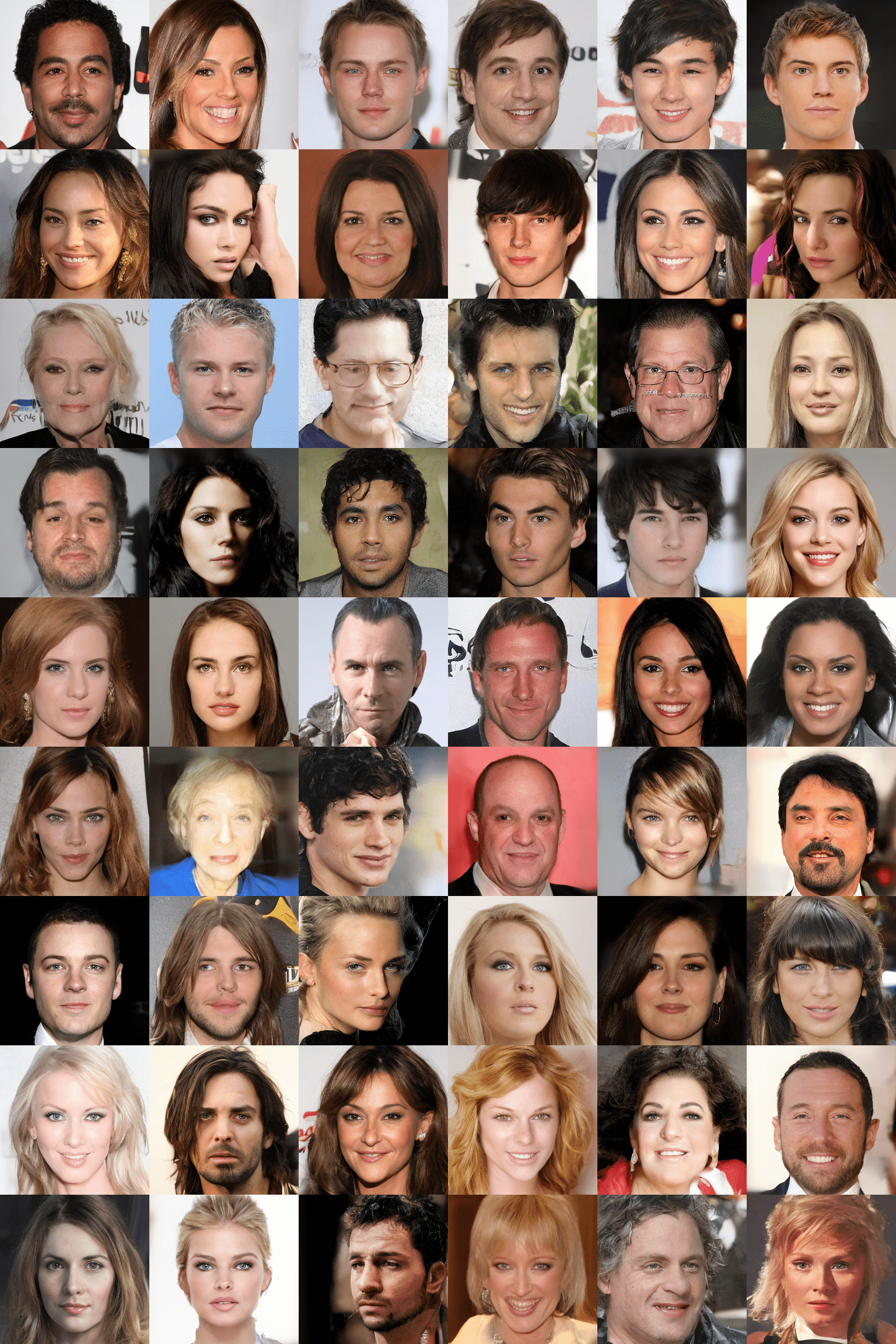}
  \caption{Randomly generated samples of CelebA-HQ using DiT-L/2+MPNN}
  \label{fig:app_CelebA_HQ_samples}
\end{figure}

\begin{figure}[ht]
  \centering
  \includegraphics[width=\linewidth]{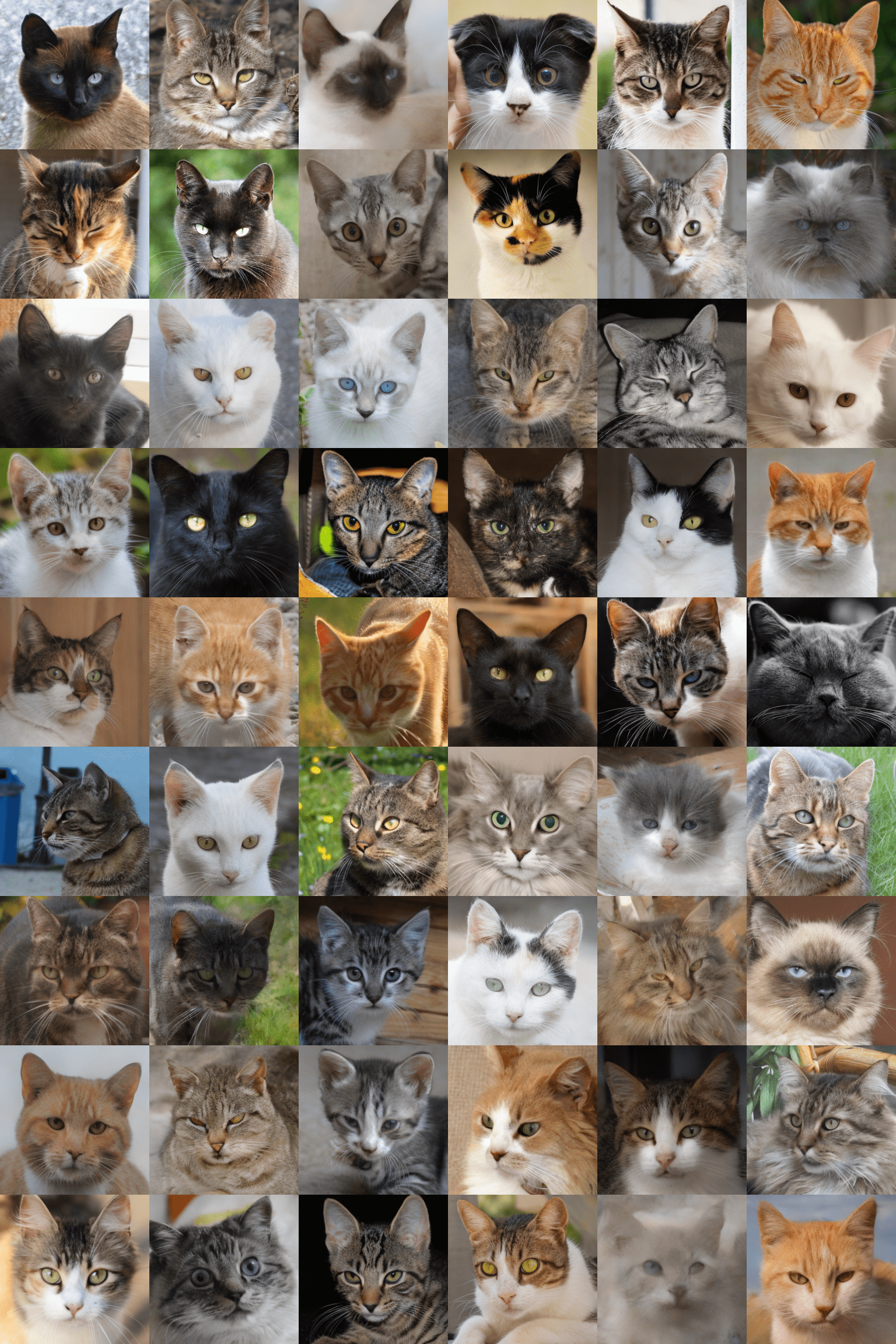}
  \caption{Randomly generated samples of AFHQ-Cat using ADM+GPS}
  \label{fig:app_AFHQ_Cat_samples}
\end{figure}

Figures~\ref{fig:app_LSUN_Church_samples}--\ref{fig:app_AFHQ_Cat_samples} show additional random samples generated by our models across the five evaluated datasets, providing readers with a broader view of typical generation quality.

\end{document}